\documentclass{article}
\usepackage{spconf,graphicx,amssymb}
\usepackage{amsmath}
\usepackage{booktabs} 
\usepackage{multirow}
\usepackage{multicol}
\usepackage[T1]{fontenc}
\usepackage{algorithm}
\usepackage{algpseudocode}

\usepackage[table,xcdraw]{xcolor}
\usepackage{setspace}
\usepackage{enumitem}
\DeclareUnicodeCharacter{2212}{-}

\algnewcommand{\LeftComment}[1]{\Statex \(\triangleright\) #1}

\title{Sem-CS: Semantic CLIPStyler for Text-Based Image Style Transfer}
\name{Chanda G Kamra$^{\star}$ \qquad Indra Deep Mastan$^{\dagger}$ \qquad Debayan Gupta$^{\star}$}  
  \address{$^{\star}$ Ashoka University, Computer Science, Sonipat, Haryana, India. \\
      $^{\dagger}$ The LNM Institute of Information Technology, Computer Science, Jaipur, Rajasthan, India.
      }

\begin{document}

\maketitle
\begin{abstract}
CLIPStyler demonstrated image style transfer with realistic textures using only the style text description (instead of requiring a reference style image). However, the ground semantics of objects in style transfer output is lost due to style spill-over on salient and background objects (content mismatch) or over-stylization. To solve this, we propose Semantic CLIPStyler (Sem-CS) that performs semantic style transfer.

Sem-CS first segments the content image into salient and non-salient objects and then transfers artistic style based on a given style text description. The semantic style transfer is achieved using global foreground loss (for salient objects) and global background loss (for non-salient objects). Our empirical results, including DISTS, NIMA and user study scores, show that our proposed framework yields superior qualitative and quantitative performance. 
\end{abstract}
%

\section{Introduction}
\label{sec:intro}
Image style transfer \cite{NEURIPS2021_df535469, li2017universal, park2019arbitrary,Li2018LearningLT, mechrez2018contextual, 8451734} aims to synthesize new images by transferring the style features such as color and texture patterns to the content image. Image style transfer can be classified into Photo-realistic style transfer \cite{BMVC2017_153, 9897202} and Artistic style transfer \cite{gatys2016image, samuth2022patch} based on the input content image and style image. One of the challenges in image style transfer comes when a user does not have the reference style image for the desired style in mind. 

\begin{figure}[!ht]
    \centering
    \begin{minipage}{0.19\linewidth}
     \centering
        \textbf{Style Text}
    \end{minipage}
    \hfill
    \begin{minipage}{0.19\linewidth}
     \centering
        \textbf{Input Image}
    \end{minipage}
    \hfill
    \begin{minipage}{0.19\linewidth}
     \centering
        CLIPStyler \cite{Kwon_2022_CVPR}
    \end{minipage}
    \hfill
    \begin{minipage}{0.19\linewidth}
     \centering
        Gen-Art\cite{genArt}
    \end{minipage}
    \hfill
    \begin{minipage}{0.19\linewidth}
     \centering
        Sem-CS (ours)
    \end{minipage}
    \fbox{
    \begin{minipage}[c][1.3cm][c]{1.3cm}
     \centering
     \small
      A monet style painting
    \end{minipage}}
        \begin{minipage}{0.19\linewidth}
         \centering
             \includegraphics[width=0.99\linewidth]{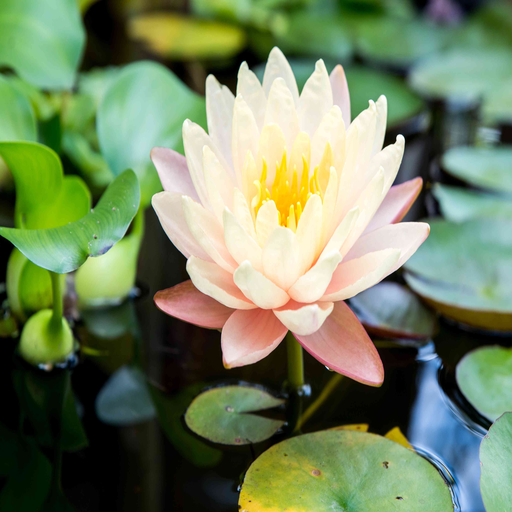}
        \end{minipage}
        \hfill
        \begin{minipage}{0.19\linewidth}
         \centering
             \includegraphics[width=0.99\linewidth]{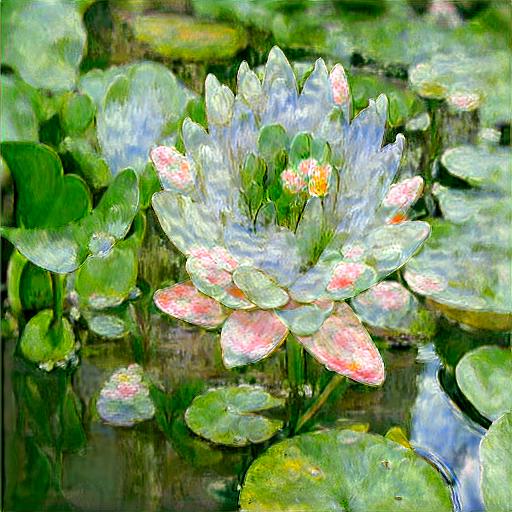}
        \end{minipage}
        \begin{minipage}{0.19\linewidth}
         \centering                 
            \includegraphics[width=0.99\linewidth]{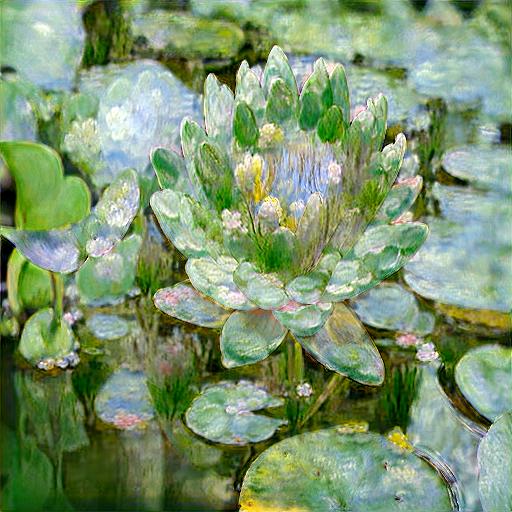}
        \end{minipage}
        \begin{minipage}{0.19\linewidth}
         \centering                 
            \includegraphics[width=0.99\linewidth]{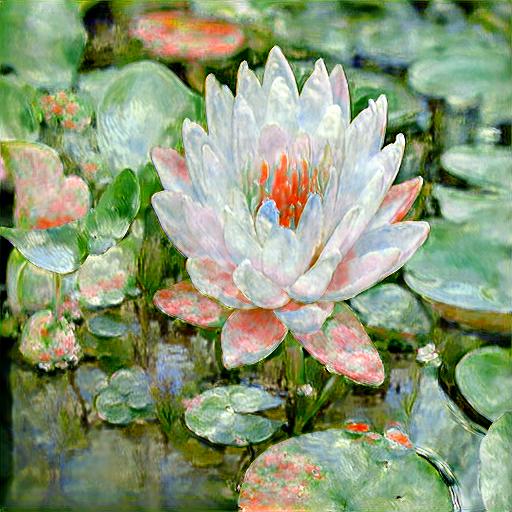}
        \end{minipage}
   \vspace{0.2cm}     
   \hrule
   \vspace{0.2cm}
    \fbox{
   \begin{minipage}[c][1.3cm][c]{1.3cm}
     \centering
     \small
      Desert Sand
    \end{minipage}}
    \hfill
    \begin{minipage}{0.19\linewidth}
     \centering
      \includegraphics[width=0.99\linewidth]{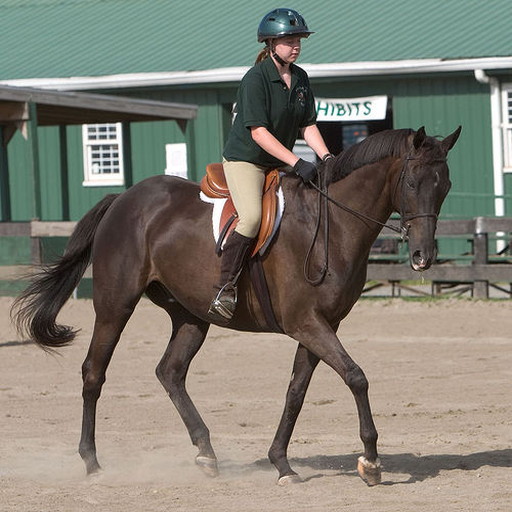}
    \end{minipage}
    \begin{minipage}{0.19\linewidth}
     \centering
      \includegraphics[width=0.99\linewidth]{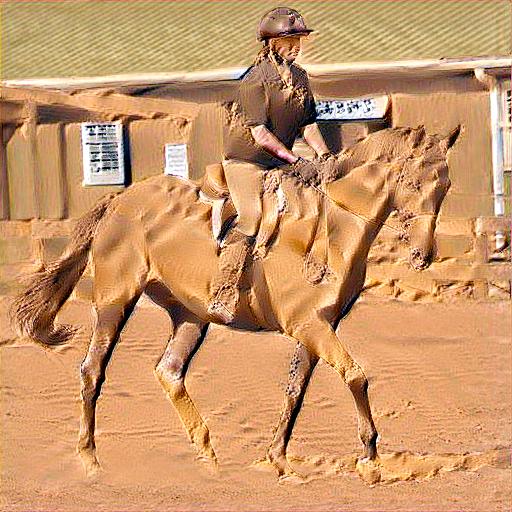}
    \end{minipage}
    \hfill
    \begin{minipage}{0.19\linewidth}
     \centering
      \includegraphics[width=0.99\linewidth]{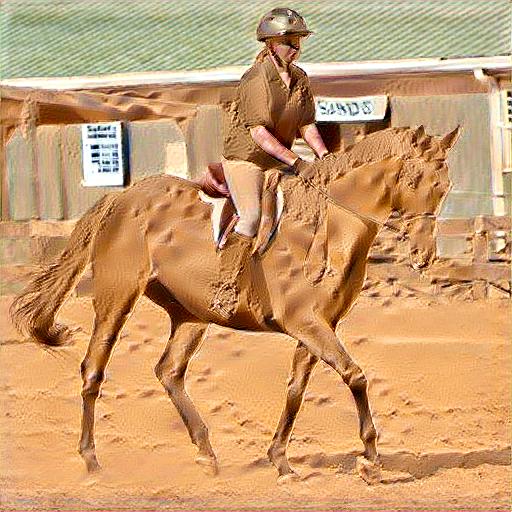}
    \end{minipage}
    \begin{minipage}{0.19\linewidth}
     \centering
      \includegraphics[width=0.99\linewidth]{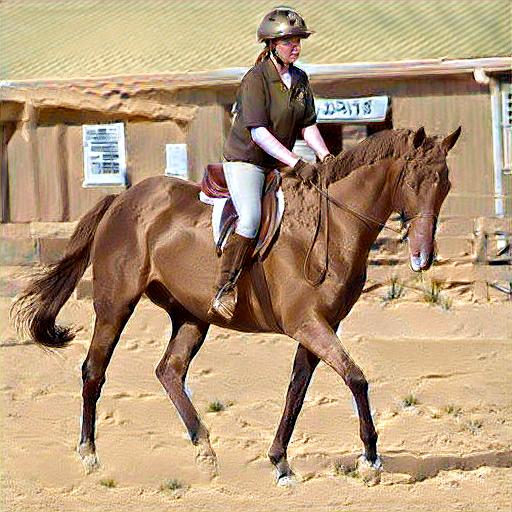}
    \end{minipage}
    \caption{\small{The figure illustrates over-stylization and the effects of content mismatch on style transfer output. \textbf{Top row} CLIPStyler\cite{Kwon_2022_CVPR} and Gen-Art\cite{genArt} over-stylize style features on salient objects and image background as the content features of the flower are lost. Sem-CS (ours) preserved the semantics of the flower. \textbf{Bottom row} CLIPStyler\cite{Kwon_2022_CVPR} and Generative Artisan\cite{genArt} outputs suffer from content mismatch as the Desert Sand style is applied to both man and horse. Sem-CS (ours) performed style transfer while minimizing content mismatch and preserves semantics.}}
    \label{fig:main_fig}
\end{figure}

Recently, CLIPStyler~\cite{Kwon_2022_CVPR} proposed a novel artistic style transfer approach that solely uses text condition to supervise style features to content images without needing a reference style image. Although CLIPStyler performs text-based style transfer, content features in the output are mostly distorted due to over-styling problem (see Fig.~\ref{fig:main_fig}-first row).

Another challenge in style transfer is when style spillover between dissimilar objects, also known as the content mismatch problem \cite{Luan2017DeepPS} (see Fig.~\ref{fig:main_fig}-first second row), occurs. The content mismatch reduces the visual quality of the style transfer output and it is inherent when the semantic objects in the style and the content features are of different types and numbers \cite{Mastan_2021_CVPR}. A good style transfer approach aims to perform style supervision by minimizing content mismatch and over-styling problem. 

    \begin{figure*}[!h]
        \centering
        \includegraphics[width=0.95\textwidth]{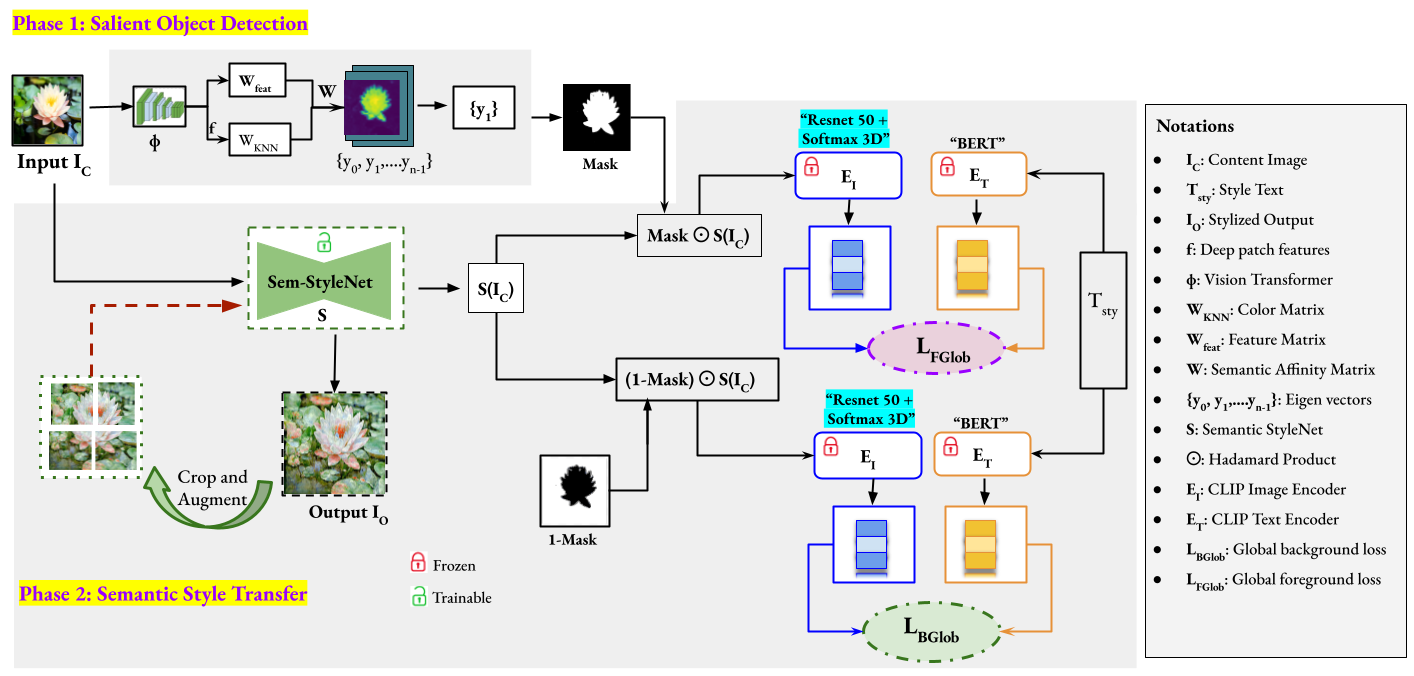}
        \vspace{-0.2cm}
    \caption{\small{The figure shows Semantic CLIPStyler (Sem-CS) framework. The two main phases Salient Object Detection and Semantic Style Transfer (shown at the top \& bottom). The proposed global foreground and background loss are illustrated on the right side. }}
        \label{fig:block1}
    \end{figure*}
    
 Generative Artisan (Gen-Art)\cite{genArt} address the over-styling problem of CLIPStyler through an FCN semantic segmentation network  \cite{long2015fully}. They controlled the degree of image style transfer in different semantic parts. However, the supervised approach to extract the semantic parts of the content image lacks generalizability to diverse content images. For example, FCN semantic segmentation network only considers 21 classes that are insufficient to represent real-world images. Also, Generative Artisan does not address the content mismatch problem inherent in the style transfer (see Fig.~\ref{fig:main_fig}).

In this paper, we propose Semantic CLIPStyler (Sem-CS), which addresses the content mismatch and over-styling problem of style transfer based on text conditions. We use the deep spectral segmentation network \cite{melas2022deep}, which extracts salient and non-salient objects of the content image in an unsupervised manner. Therefore, Sem-CS (ours) is generalizable to various real-world images. 

Sem-CS applies style description on salient or non-salient objects based on the text conditions to prevent over-stylization and content mismatch of style transfer output. The key idea is to perform semantic style transfer using the proposed global foreground and background loss. Sem-CS also archives controllable generation in transferring texture information for multiple text conditions. The major contributions of the proposed Sem-CS framework are described as follows.
    \begin{itemize}[noitemsep,leftmargin=*]
        \item We proposed a novel Sem-CS framework to perform style transfer with text condition (Algorithm~\ref{alg:cap}). 
        \item We proposed global foreground and global background loss to supervise style features semantically on the output (Sec.~\ref{sec:approach}).  
        \item We provide reference based quality assessment using DISTS \cite{ding2020image} and no-reference quality assessment using NIMA \cite{talebi2018nima} to show Sem-CS outperforms baselines (Table~\ref{table:nima_dist}). 
    \end{itemize}
\vspace{-0.3cm}
     \begin{algorithm}[!h]
    \setstretch{1.08}
    \caption{Semantic CLIPStyler framework.}
    \label{alg:cap}
    \begin{algorithmic}[1] \vspace{3pt}
    \State \textbf{\textsc{Sem-CS}($E_T$, $E_I$, $I_C$, $T_{sty}$, $\phi$, $S$)}
    \LeftComment {{\color{gray} \textit{Compute Mask for salient objects identification}}} 
    \State \hspace{0.5cm}  $W$ = AffinityMatrix($I_c$, $\phi$, )
    \State \hspace{0.5cm}  $\{y_0$, $y_1$, $\ldots$, $y_{n-1}\}$ = Eigen\_Decomposition($W$)  
    \State \hspace{0.5cm}  $Mask$ =  Extract\_Salient\_Object($y_1$)  \vspace{4pt}
    \LeftComment {{\color{gray} \textit{Perform Semantic Style Transfer}}} 
    \State\hspace{0.5cm} $t_{fg}$, $t_{bg}$ = Parse\_Style\_Text($T_{sty}$)
    \State\hspace{0.5cm} $I_{fg}$, $I_{bg}$ = $Mask \odot S(I_C), (1-Mask) \odot S(I_C)$
    \LeftComment {{\color{teal} \textit{~~~~Global Foreground Loss}}}
    \State \hspace{0.5cm}  Compute Foreground Image Direction Loss $\Delta fg_{I}$  
    \State \hspace{0.5cm}  Compute Foreground Text Direction Loss $\Delta fg_{T}$
    \State \hspace{0.5cm} $\mathcal{L}_{FGlob}$ = Cosine\_similarity($\Delta fg_{I}, \Delta fg_{T}$)
    \LeftComment {{\color{teal} \textit{~~~~Global Background Loss}}}
    \State \hspace{0.5cm}  Compute Background Image Direction Loss $\Delta bg_{I}$ 
    \State \hspace{0.5cm} Compute Background Text Direction Loss $\Delta bg_{T}$
    \State \hspace{0.5cm} $\mathcal{L}_{BGlob}$ = Cosine\_similarity($\Delta bg_{I}$, $\Delta bg_{T}$)
    \LeftComment {{\color{teal} \textit{~~~~Minimize loss and compute output $I_O$}}}
    \State \hspace{0.5cm}  $I_O = \displaystyle\min_{{\theta_S}} \big( \mathcal{L}_{FGlob} + \lambda_{bg} \mathcal{L}_{BGlob} \big)$ 
\end{algorithmic}
\end{algorithm}

  \begin{figure*}[!h]
        \begin{minipage}{0.10\textwidth}
         \centering
            \small
            Input Image
        \end{minipage}
        \begin{minipage}{0.10\textwidth}
         \centering
         \small
            Style Text
        \end{minipage}
        \hfill
        \begin{minipage}{0.09\textwidth}
         \centering
         \small
            CS\cite{Kwon_2022_CVPR}
        \end{minipage}
        \begin{minipage}{0.09\textwidth}
         \centering
         \small
            Gen-Art\cite{genArt}
        \end{minipage}
        \begin{minipage}{0.09\textwidth}
         \centering
         \small
            Sem-CS (ours)
        \end{minipage} 
        \hfill
        \hfill
        \begin{minipage}{0.10\textwidth}
         \centering
         \small
            Input Image
        \end{minipage}
        \begin{minipage}{0.10\textwidth}
         \centering
         \small
            Style text
        \end{minipage}
        \hfill
        \begin{minipage}{0.09\textwidth}
         \centering
         \small
            CS \cite{Kwon_2022_CVPR}
        \end{minipage}
        \begin{minipage}{0.09\textwidth}
         \centering
         \small
            Gen-Art\cite{genArt}
        \end{minipage}
        \begin{minipage}{0.09\textwidth}
         \centering
         \small
            Sem-CS (ours)
        \end{minipage} 

       \begin{minipage}{0.09\textwidth}
         \centering
             \includegraphics[width=0.99\linewidth]{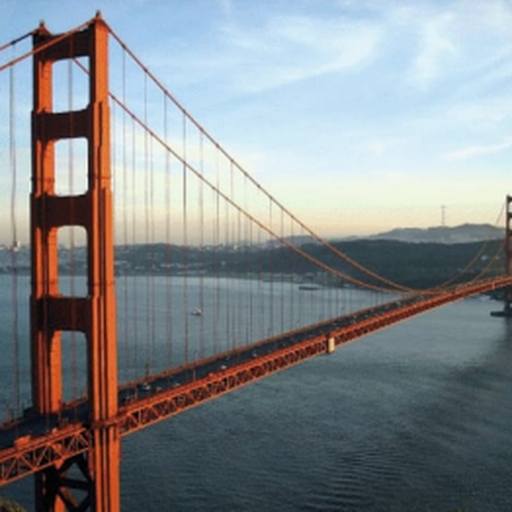}
        \end{minipage}
         \fbox{\begin{minipage}[c][1.35cm][c]{1.5cm}
         \centering
         \footnotesize
            Acrylic painting
        \end{minipage}}
        \hfill
        \begin{minipage}{0.09\textwidth}
         \centering
             \includegraphics[width=0.99\linewidth]{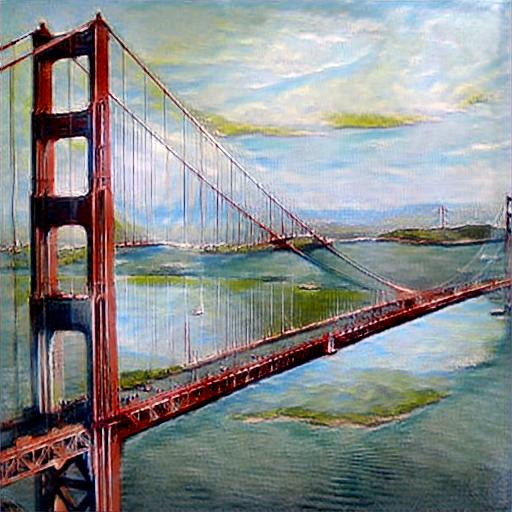}
        \end{minipage}
        \begin{minipage}{0.09\textwidth}
         \centering                 
            \includegraphics[width=0.99\linewidth]{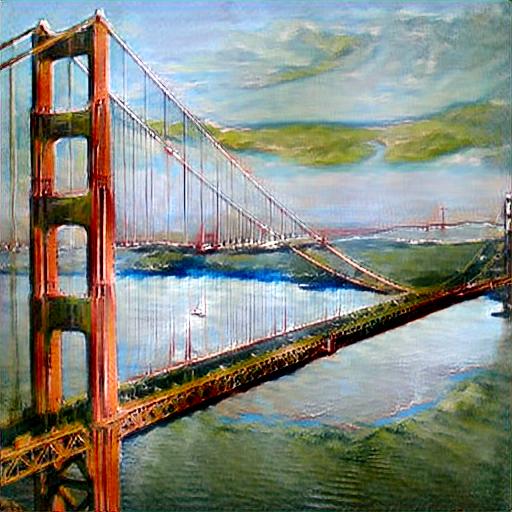}
        \end{minipage}
        \begin{minipage}{0.09\textwidth}
         \centering                 
            \includegraphics[width=0.99\linewidth]{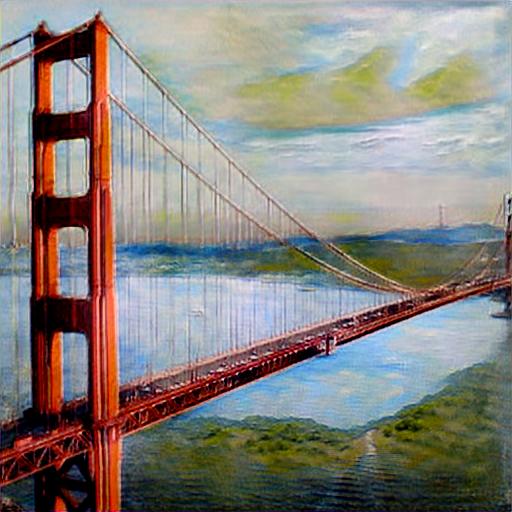}
        \end{minipage}
        \hfill
        \textbf{\vline}
        \hfill
    \begin{minipage}{0.09\textwidth}
         \centering             \includegraphics[width=0.99\linewidth]{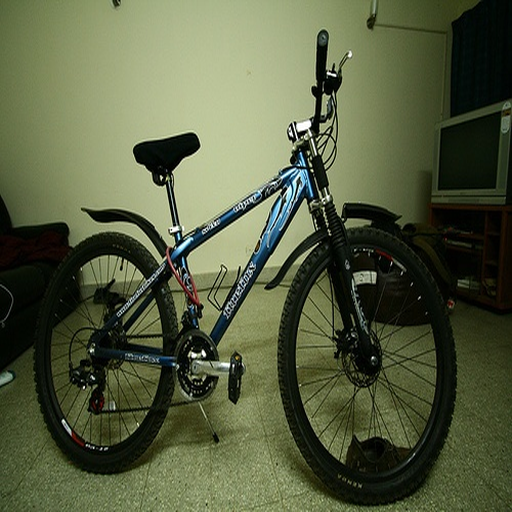}
        \end{minipage}
        \fbox{\begin{minipage}[c][1.35cm][c]{1.5cm}
         \centering
         \footnotesize
            Snowy
        \end{minipage}}
        \hfill
        \begin{minipage}{0.09\textwidth}
         \centering
             \includegraphics[width=0.99\linewidth]{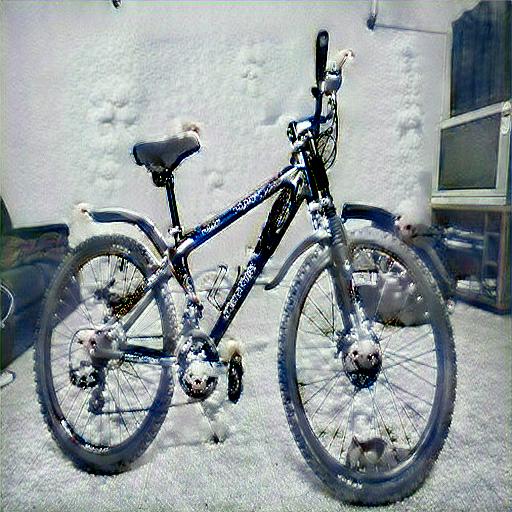}
        \end{minipage}
        \begin{minipage}{0.09\textwidth}
         \centering                 
            \includegraphics[width=0.99\linewidth]{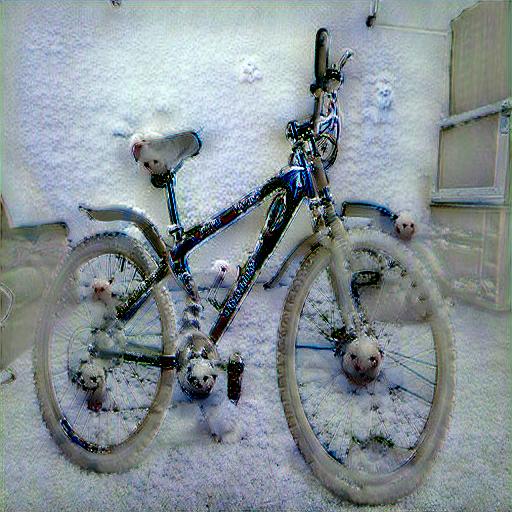}
        \end{minipage}
        \begin{minipage}{0.09\textwidth}
         \centering                 
            \includegraphics[width=0.99\linewidth]{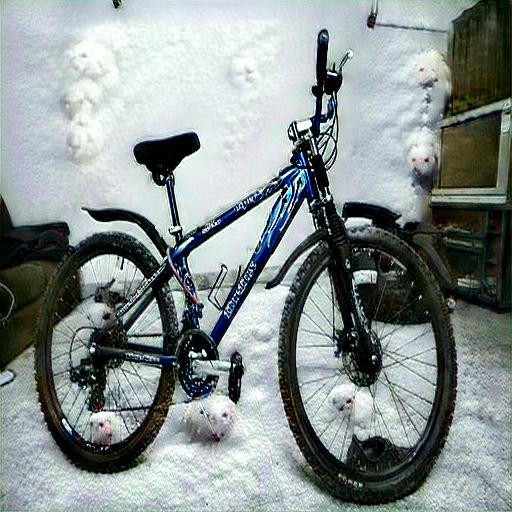}
        \end{minipage}
  
    \begin{minipage}{0.09\textwidth}
         \centering
             \includegraphics[width=0.99\linewidth]{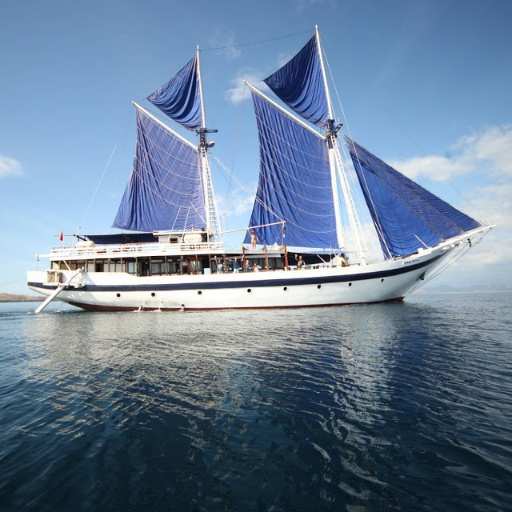}
        \end{minipage}
       \fbox{\begin{minipage}[c][1.35cm][c]{1.5cm}
         \centering
         \footnotesize
            A graffiti style painting
        \end{minipage}}
        \hfill
        \begin{minipage}{0.09\textwidth}
         \centering             \includegraphics[width=0.99\linewidth]{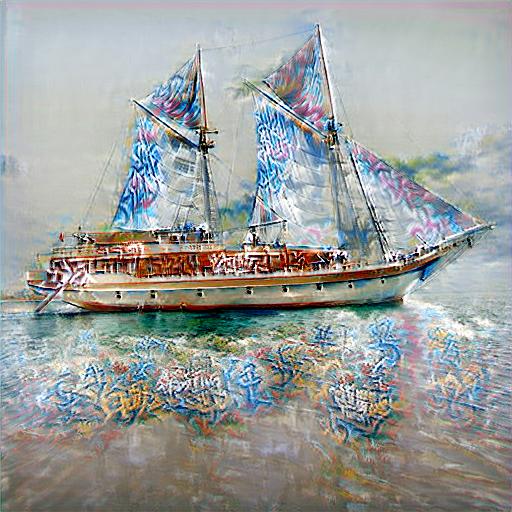}
        \end{minipage}
        \begin{minipage}{0.09\textwidth}
         \centering                 
            \includegraphics[width=0.99\linewidth]{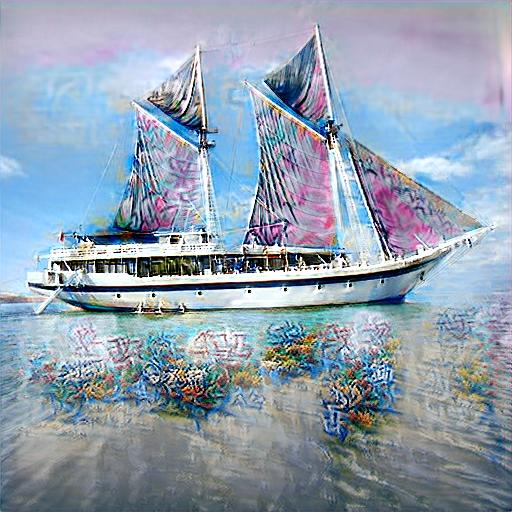}
        \end{minipage}
        \begin{minipage}{0.09\textwidth}
         \centering             \includegraphics[width=0.99\linewidth]{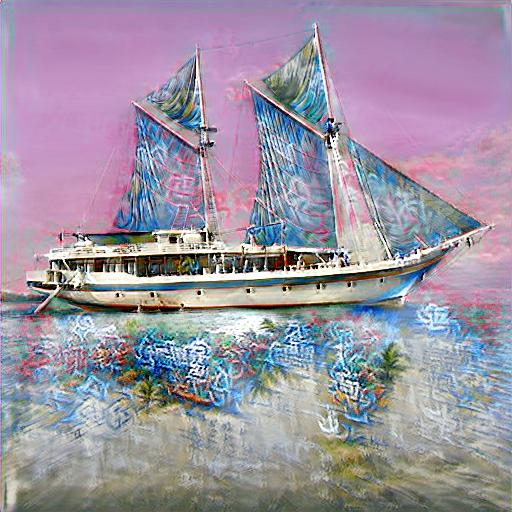}
        \end{minipage}
        \hfill
        \textbf{\vline}
        \hfill
        \begin{minipage}{0.09\textwidth}
         \centering
             \includegraphics[width=0.99\linewidth]{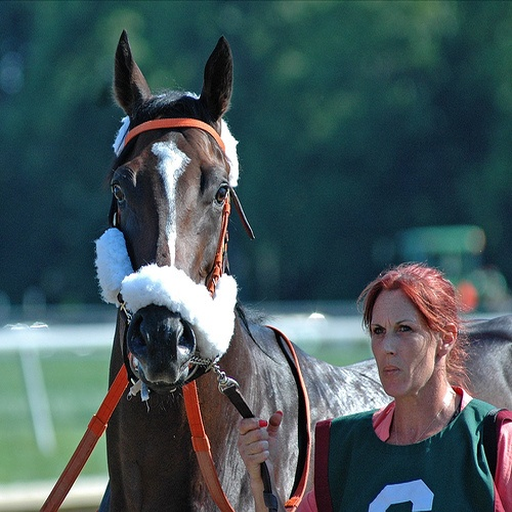}
        \end{minipage}
        \fbox{\begin{minipage}[c][1.35cm][c]{1.5cm}
         \centering
         \footnotesize
            Red rocks
        \end{minipage}}
        \hfill
        \begin{minipage}{0.09\textwidth}
         \centering
             \includegraphics[width=0.99\linewidth]{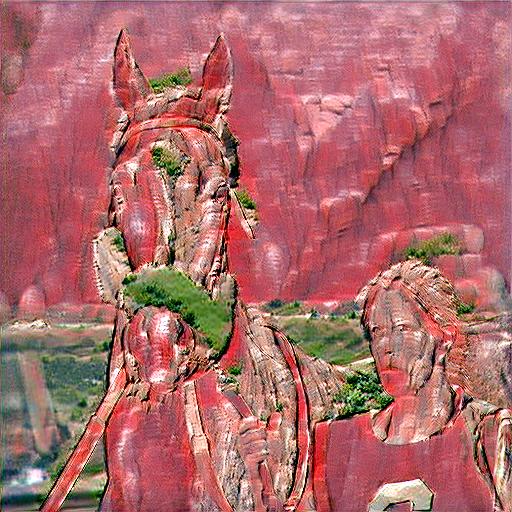}
        \end{minipage}
        \begin{minipage}{0.09\textwidth}
         \centering                 
            \includegraphics[width=0.99\linewidth]{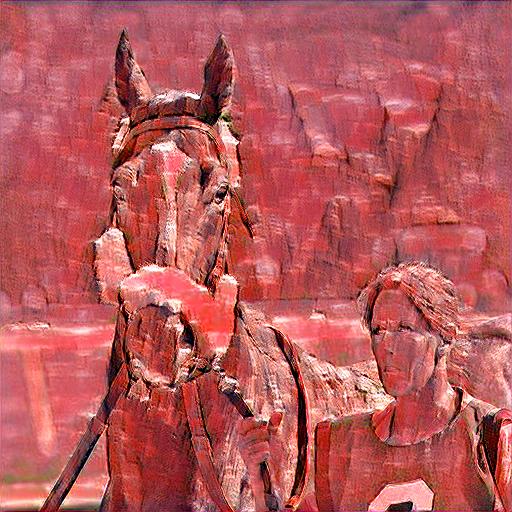}
        \end{minipage}
        \begin{minipage}{0.09\textwidth}
         \centering                 
            \includegraphics[width=0.99\linewidth]{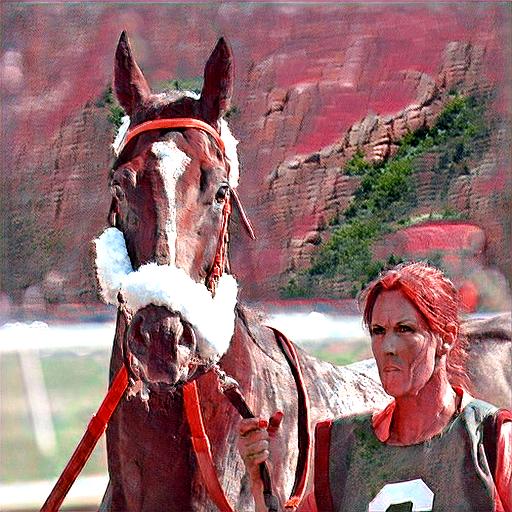}
        \end{minipage}

                \begin{minipage}{0.09\textwidth}
         \centering
             \includegraphics[width=0.99\linewidth]{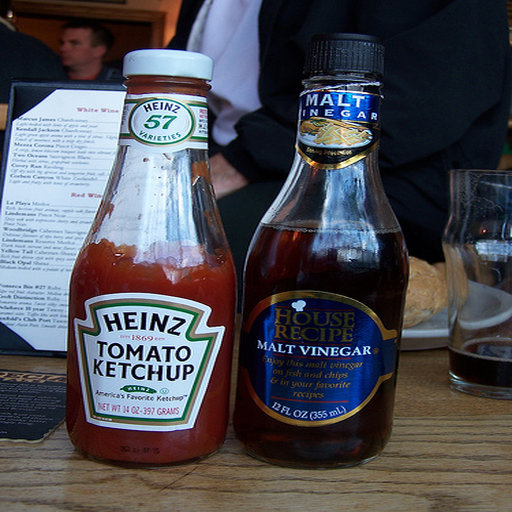}
        \end{minipage}
        \fbox{\begin{minipage}[c][1.35cm][c]{1.5cm}
         \centering
         \footnotesize
            A fauvism style painting
        \end{minipage}}
        \hfill
        \begin{minipage}{0.09\textwidth}
         \centering             \includegraphics[width=0.99\linewidth]{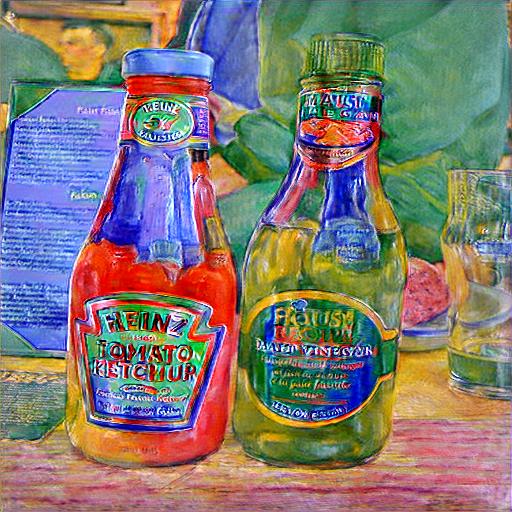}
        \end{minipage}
        \begin{minipage}{0.09\textwidth}
         \centering            
        \includegraphics[width=0.99\linewidth]{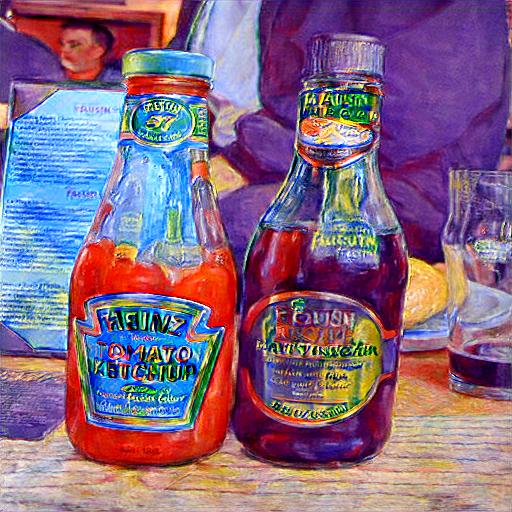}
        \end{minipage}
        \begin{minipage}{0.09\textwidth}
         \centering                 
        \includegraphics[width=0.99\linewidth]{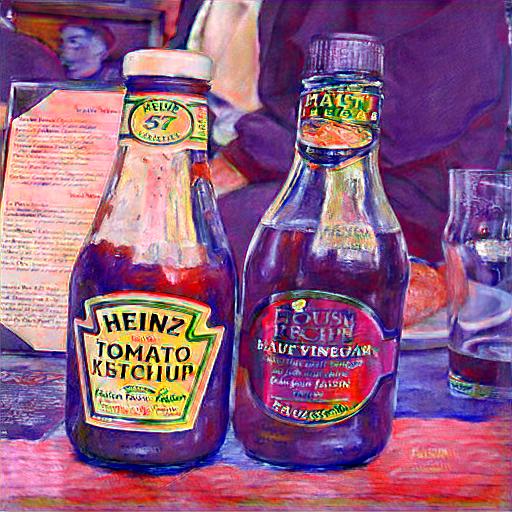}
        \end{minipage}
        \hfill
        \textbf{\vline}
        \hfill
        \begin{minipage}{0.09\textwidth}
         \centering
             \includegraphics[width=0.99\linewidth]{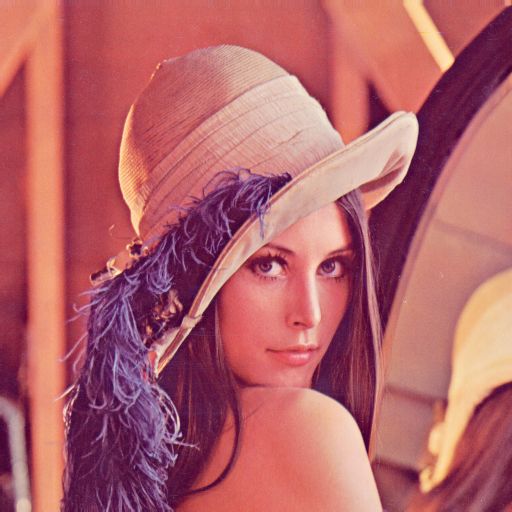}
        \end{minipage}
        \fbox{\begin{minipage}[c][1.35cm][c]{1.55cm}
         \centering
         \footnotesize
            A watercolor painting with purple brush
        \end{minipage}}
        \hfill
        \begin{minipage}{0.09\textwidth}
         \centering
             \includegraphics[width=0.99\linewidth]{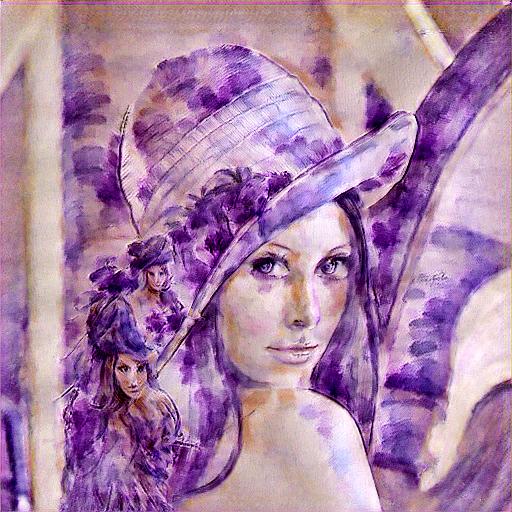}
        \end{minipage}
        \begin{minipage}{0.09\textwidth}
         \centering                 
            \includegraphics[width=0.99\linewidth]{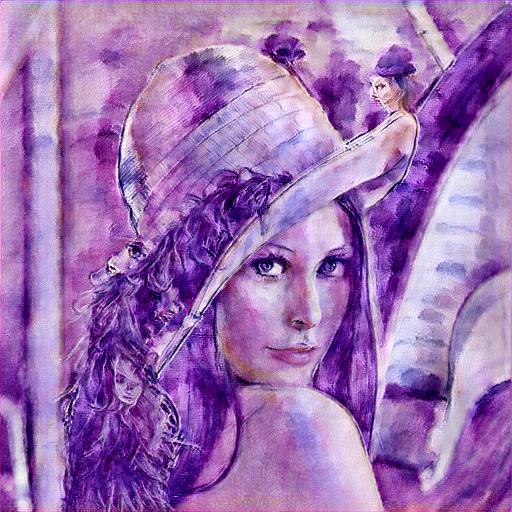}
        \end{minipage}
        \begin{minipage}{0.09\textwidth}
         \centering                 
            \includegraphics[width=0.99\linewidth]{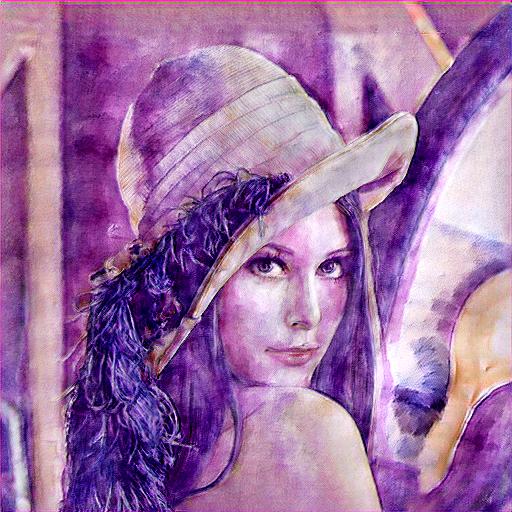}
        \end{minipage}
                \begin{minipage}{0.09\textwidth}
         \centering
             \includegraphics[width=0.99\linewidth]{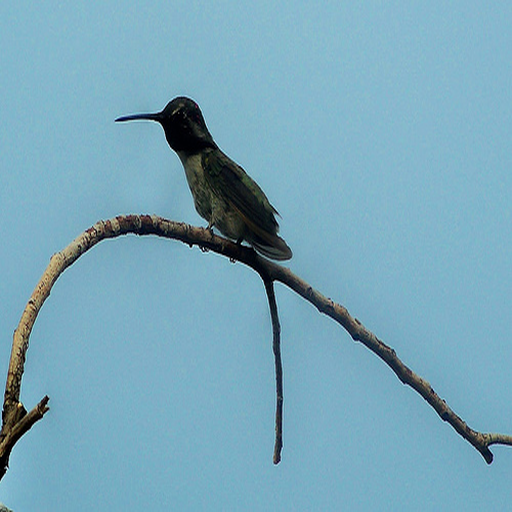}
        \end{minipage}
        \fbox{\begin{minipage}[c][1.35cm][c]{1.5cm}
         \centering
         \footnotesize
            An oil painting of white roses
        \end{minipage}}
        \hfill
        \begin{minipage}{0.09\textwidth}
         \centering             \includegraphics[width=0.99\linewidth]{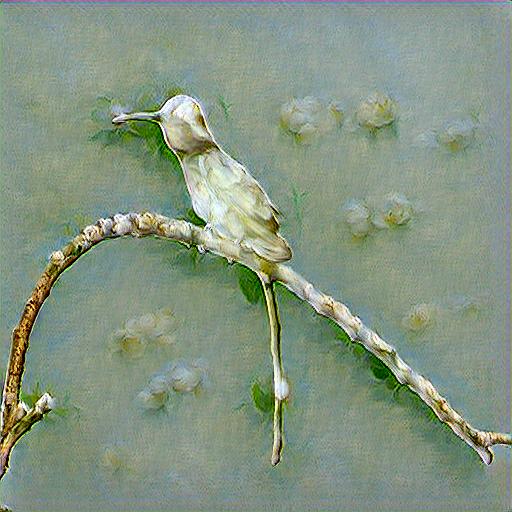}
        \end{minipage}
        \begin{minipage}{0.09\textwidth}
         \centering            
        \includegraphics[width=0.99\linewidth]{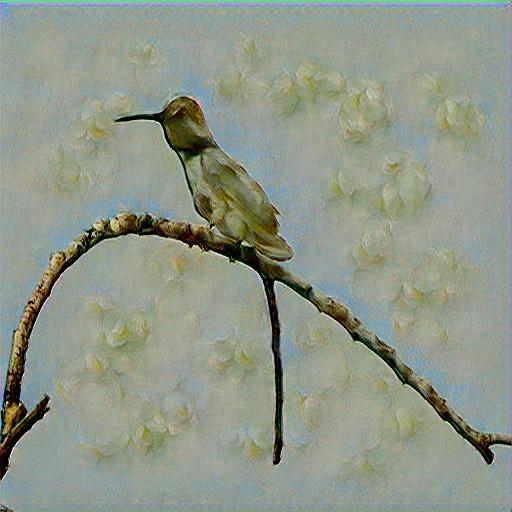}
        \end{minipage}
        \begin{minipage}{0.09\textwidth}
         \centering                 
        \includegraphics[width=0.99\linewidth]{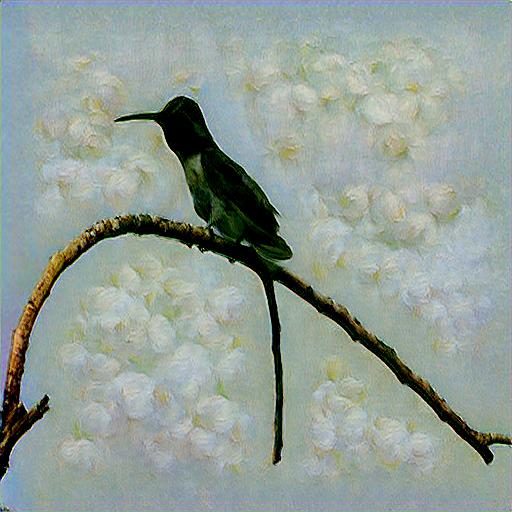}
        \end{minipage}
        \hfill
        \textbf{\vline}
        \hfill
        \begin{minipage}{0.09\textwidth}
         \centering
             \includegraphics[width=0.99\linewidth]{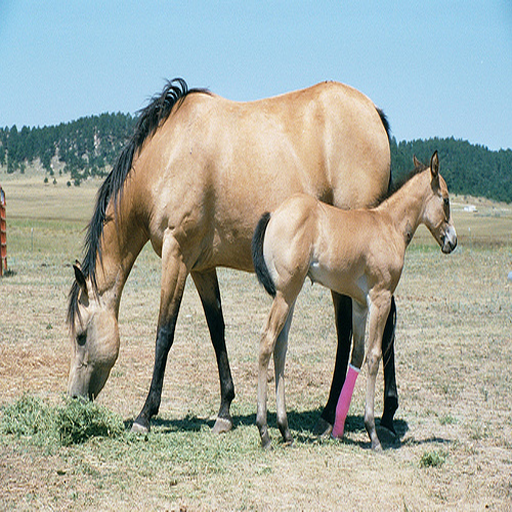}
        \end{minipage}
        \fbox{\begin{minipage}[c][1.35cm][c]{1.55cm}
         \centering
         \footnotesize
            A watercolor painting of leaf
        \end{minipage}}
        \hfill
        \begin{minipage}{0.09\textwidth}
         \centering
             \includegraphics[width=0.99\linewidth]{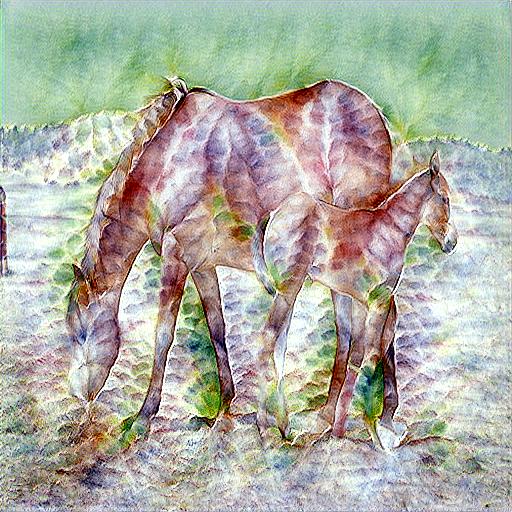}
        \end{minipage}
        \begin{minipage}{0.09\textwidth}
         \centering                 
            \includegraphics[width=0.99\linewidth]{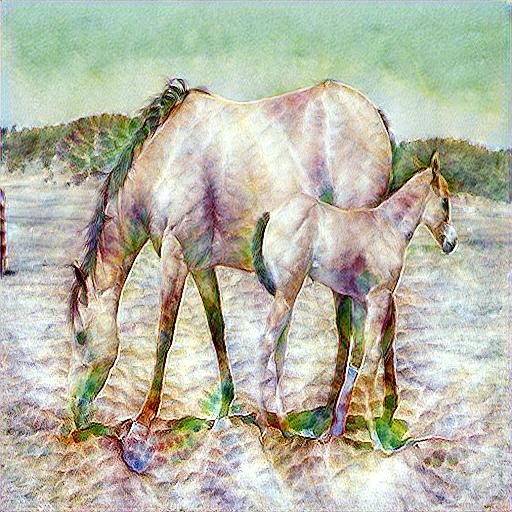}
        \end{minipage}
        \begin{minipage}{0.09\textwidth}
         \centering              
    \includegraphics[width=0.99\linewidth]{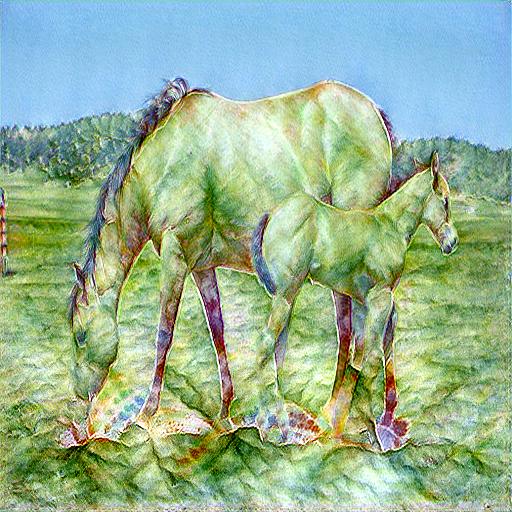}
        \end{minipage}
    \label{fig:single_style_res}
    \vspace{-0.1cm}
    \caption{\small{Visual comparison of image style transfer outputs with single text condition a) Left solve the over-stylization problem. b) Right solves the content mismatch problem. {\color{blue} (images are best viewed after zooming)}.} Note that CS = CLIPStyler.} 
\end{figure*}   
\section{Our Method}
\label{sec:approach}
This section describes proposed Semantic CLIPStyler (Sem-CS) framework. It has two major phases: Salient Object Detection and Semantic Style Transfer (Fig.~\ref{fig:block1} and Algorithm~\ref{alg:cap}). \vspace{2pt}

     \noindent \textit{\textbf{Salient Object Detection:}} In the first phase, we compute the masks for salient objects present in the content image  (Algorithm~\ref{alg:cap}, lines 2-4). The mask for salient objects is computed using three steps. First, we compute the affinity matrix (W) of content image $I_C$ from the attention block of the last layers of feature extractor $\phi$. Secondly, we find the eigenvectors of the laplacian of the affinity matrix. Finally, in an unsupervised setting, we extracts the mask from the eigenvector $y_1$. \vspace{2pt}
     
     \noindent \textit{\textbf{Semantic Style Transfer:}} In the second phase, we use Sem-StyleNet $S$ to transfer style features to the salient objects and background objects based on the text conditions (Fig.~\ref{fig:block1}). To make the stylized output more robust, we used ResNet50 with softmax3d \cite{wang2021rethinking} for the image encoder. We propose \textit{global foreground loss} and \textit{global background loss} to perform semantic style transfer. We describe these loss function as follows. \vspace{2pt}


    \noindent \textbf{\textit{Global Foreground Loss}.} It ensures that relevant style text applies to the salient objects present in the output. To maintain the diversity of generated stylized outputs directional CLIP loss \cite{gal2022stylegan} is computed instead of global CLIP loss\cite{patashnik2021styleclip} by aligning the CLIP-space direction between the text-image pairs of input and output. Foreground text directional loss  $(\Delta fg_{T})$ is defined to be the difference between source text embedding $(t_{src})$ and foreground style text embedding $(t_{fg})$ as described in Eq.~\ref{eq:fg_t}.   
         \begin{equation}
            \label{eq:fg_t}
            \Delta fg_{T} = E_{T}(t_{fg}) - E_{T}(t_{src})
    \end{equation}
    Here, $E_{T}$ is the CLIP text-encoder and $t_{src}$ is set to "Photo". Foreground image directional loss $(\Delta fg_{I})$ is computed between the embedding of salient objects and style transfer output. Given the content image $I_C$ and $Mask$, hadamard product $\odot$ is taken between them to extract features for salient objects as $I_{fg} = Mask \odot (S(I_C))$. Next, $\Delta fg_{I}$ is computed as follows.
        \begin{equation}
        \label{eq:fg_i}
        \Delta fg_{I} = E_{I}(I_{fg}) - E_I(I_C)
    \end{equation}
    Here, $E_{I}$ is the CLIP image encoder. Finally, Global foreground loss $(\mathcal{L}_{FGlob})$ is computed by taking cosine similarity between CLIP-Space direction of foreground of image and style texts. 
    \begin{equation}
        \label{eq:fg_glob}
        \mathcal{L}_{FGlob} = 1- \frac{\Delta fg_{I} . \Delta fg_{T}}{\lvert{\Delta fg_{I}}\rvert \lvert{\Delta fg_{T}}\rvert}
    \end{equation}
    Here, one minus the cosine similarity is taken between the image and text directional loss which represents the distance and we aim to minimize it. In other words, the global foreground loss minimizes the distance between the image direction loss and text direction loss for salient objects. \vspace{2pt}

    \noindent \textbf{\textit{Global Background Loss}.} It is computed for style features supervision of output image background. Similar to global foreground loss, we compute background text directional loss $(\Delta bg_{T})$ for style background as given in Eq.~\ref{eq:bg_T}.
         \begin{equation}
         \label{eq:bg_T}
            \Delta bg_{T} = E_{T}(t_{bg}) - E_{T}(t_{src})
    \end{equation}
    Here, $t_{bg}$ is the style text condition for the background. Also, background image directional loss $\Delta bg_{I}$ is computed as shown in Eq.~\ref{eq:bg_i}. We take Hadamard product between the background mask and generated image $I_{bg} = (1-Mask) \odot S(I_C)$ to extract background features. Next, $\Delta bg_{I}$ is computed as below.
       \vspace{-0.2cm}
        \begin{equation}
        \label{eq:bg_i}
        \Delta bg_{I} = E_{I}(I_{bg}) - E_I(I_C)
    \end{equation}
    Finally, global background loss $\mathcal{L}_{BGlob}$ is computed to minimize the distance between image and text directional losses for background objects as described in Eq.~\ref{eq:bg_loss}.
    \begin{equation}
    \label{eq:bg_loss}
        \mathcal{L}_{BGlob} = 1- \frac{\Delta bg_{I} . \Delta bg_{T}}{\lvert{\Delta bg_{I}}\rvert \lvert{\Delta bg_{T}}\rvert}
    \end{equation}   
    Here, global background loss $\mathcal{L}_{BGlob}$ helps to perform controllable style transfer for background objects in the style transfer outputs. \vspace{2pt}

    \noindent \textbf{\textit{Other Loss}.} We also use content loss and a total variation regularization loss \cite{Kwon_2022_CVPR}.

\section{Experimental Results}
\label{ssec:subheadresult}

This section illustrates results for text-based image style transfer qualitatively and quantitatively. Sem-CS (ours) preserve the semantics of objects in the outputs images while minimizing over-stylization and content mismatch. For example, in Fig.~3, first row left side, CLIPStyler and Generative Artisan outputs are over-stylized (Acrylic style is spilled below the bridge and onto sky) as the content features of the water are lost. Sem-CS (ours) preserved the semantics of the bridge. Similarly, in the first row, on the right side, CLIPStyler and Generative Artisan outputs suffer from content mismatch as the Snowy style is applied to bicycle and background. Sem-CS performed style transfer while minimizing content mismatch effects of Snowy style feature
       \begin{figure}[!htb]
       \centering
        \begin{minipage}{0.25\linewidth}
         \centering
         \small
            \textbf{Input Image}
        \end{minipage}
      \begin{minipage}{0.25\linewidth}
         \centering
         \small
            Gen-Art \cite{genArt}
        \end{minipage}
      \begin{minipage}{0.25\linewidth}
         \centering
         \small
            Sem-CS (ours)
        \end{minipage}
    \begin{minipage}{0.25\linewidth}
         \centering
             \includegraphics[width=0.99\linewidth]{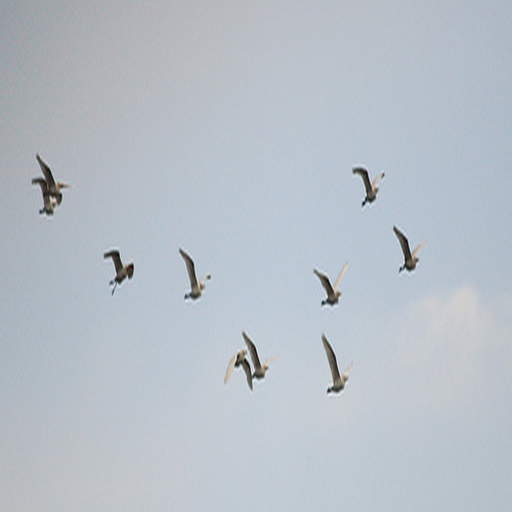}
        \end{minipage}
        \begin{minipage}{0.25\linewidth}
         \centering
             \includegraphics[width=0.99\linewidth]{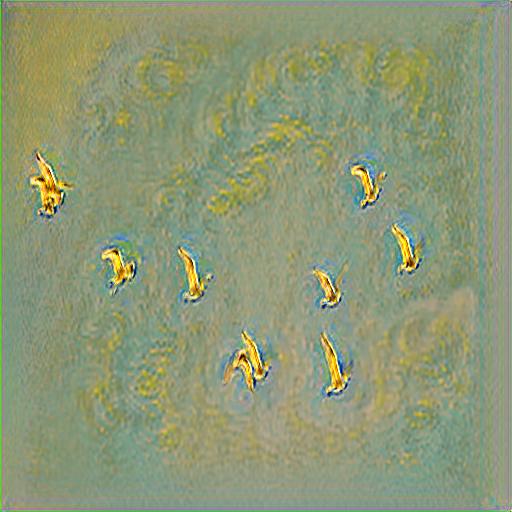}
        \end{minipage}
        \begin{minipage}{0.25\linewidth}
         \centering
             \includegraphics[width=0.99\linewidth]{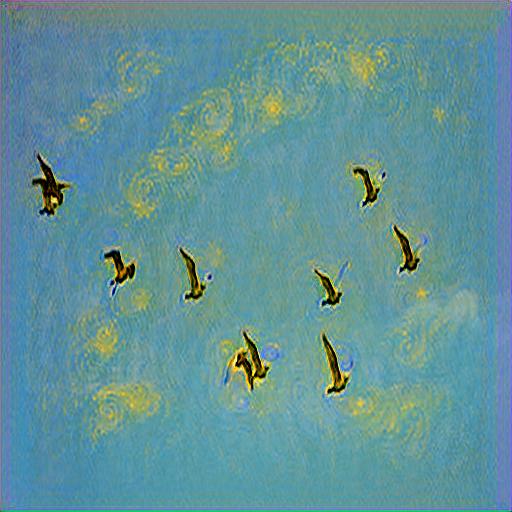}
        \end{minipage}
        
        \begin{minipage}{0.25\linewidth}
         \centering
             \includegraphics[width=0.99\linewidth]{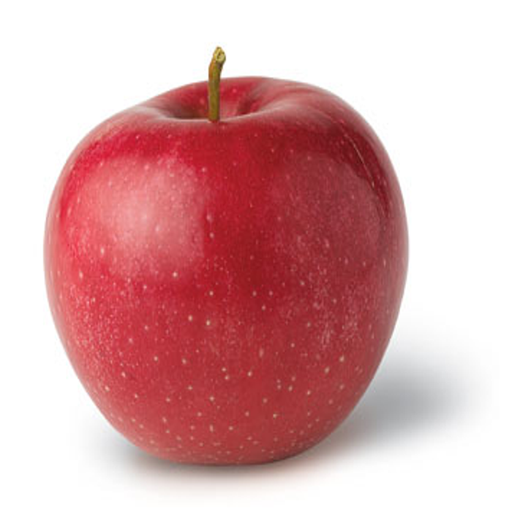}
        \end{minipage}
        \begin{minipage}{0.25\linewidth}
         \centering
             \includegraphics[width=0.99\linewidth]{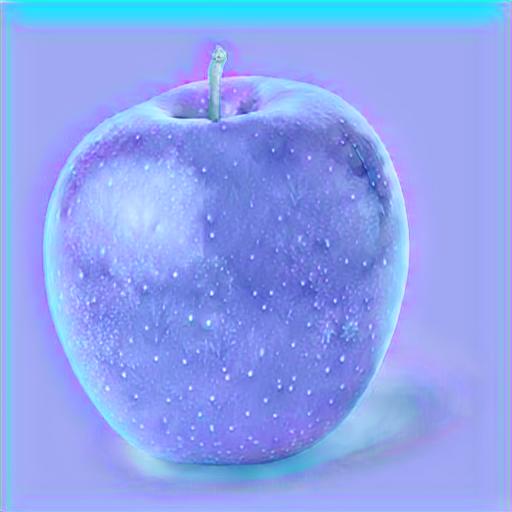}
        \end{minipage}
        \begin{minipage}{0.25\linewidth}
         \centering
             \includegraphics[width=0.99\linewidth]{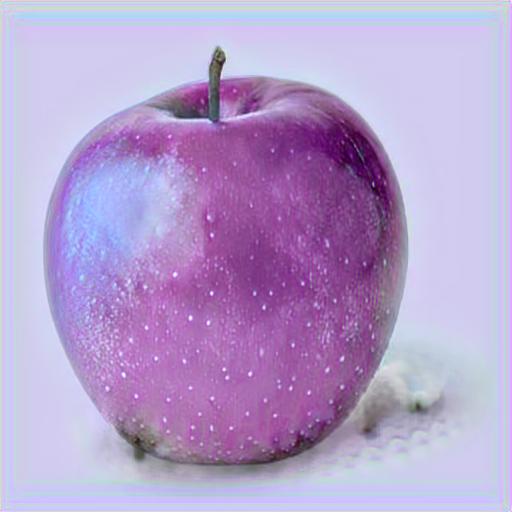}
        \end{minipage}
    \vspace{-0.1cm}
        \caption{\small{Stylized outputs with two text conditions (foreground \textbf{F} and  background \textbf{B}). a) Top Row: \textbf{F}: Pop Art; \textbf{B}: Starry Night by Vincent Van Gogh. b) Bottom Row: \textbf{F}: Red Rocks; \textbf{B}: Snowy.}}
        \label{fig:double_style_res}
    \end{figure}
    
        \vspace{2pt}
        \noindent \textbf{\textit{Quantitative Results.}}
        We evaluated Sem-CS framework with  NIMA \cite{talebi2018nima} and  DISTS \cite{ding2020image} scores. NIMA \cite{talebi2018nima} is a no-reference-based image quality metric that effectively predicts the quality of distribution ratings with a significant correlation to ground truth ratings. DISTS \cite{ding2020image} is a full reference image quality metric that captures the texture transfer and structure similarity in the style transfer output.  Table~\ref{table:nima_dist} reports the average scores of top-100 output images. It could be observed that Sem-CS scores are higher than the baseline methods CLIPStyler and Generative Artisan, respectively. This justifies that adding global foreground and background losses improve the image quality of stylized output. Sem-CS minimize content mismatch and prevents distortion of objects present in output image when supervising various style features. 

    \vspace{2pt}
    \noindent \textbf{\textit{User Study.}}
     We also conducted user studies to validate preserved semantics of objects while transferring the style texts onto content image (Table~\ref{table:nima_dist}). We randomly sample 5 groups of 15 images from the outputs produced above with 10 images from single style texts stylized outputs and five images from double style texts stylized outputs. All 5 $\times$ 15 stylized outputs were distributed anonymously and randomly to 40 participants. They were asked to observe the stylized results from different methods and choose the image that looks better in quality and matches the style text. Table~\ref{table:nima_dist} shows that Sem-CS outperforms  baseline methods.\vspace{2pt}

\noindent \textbf{\textit{Ablation Studies.}} Fig.~\ref{fig:double_style_res} illustrate ablation studies for style transfer using double style texts condition. The double-style texts is challenging due to style supervision is required for salient and background objects. Therefore, double-style texts requires more controllable generation capabilities for style transfer. We evaluated Sem-CS framework with  NIMA \cite{talebi2018nima} and  DISTS \cite{ding2020image} scores on 100 stylized outputs. Table~\ref{table:ablation} describes that Sem-CS outperforms Generative Artisan \cite{genArt}. Also, note that the user study scores of style transfer outputs for double text condition for Sem-CS are higher. 

\begin{table}[!ht]\small{
\begin{center}\renewcommand{\arraystretch}{1.25}
\begin{tabular}{cccc}
                                         & \cellcolor[HTML]{ECF4FF}CLIPStyler & \cellcolor[HTML]{EFEFEF}Gen-Art & \cellcolor[HTML]{FFFFC7}Sem-CS (ours) \\ \cline{2-4} 
\multicolumn{1}{l|}{\textit{DISTS}}      & \multicolumn{1}{c|}{0.32}              & \multicolumn{1}{c|}{0.25}                      & \multicolumn{1}{c|}{\textbf{0.34}}                 \\ \cline{2-4} 
\multicolumn{1}{l|}{\textit{NIMA}}       & \multicolumn{1}{c|}{4.61}              & \multicolumn{1}{c|}{4.34}                      & \multicolumn{1}{c|}{\textbf{5.34}}                 \\ \cline{2-4} 
\multicolumn{1}{l|}{\textit{User Study}} & \multicolumn{1}{c|}{28.3}              & \multicolumn{1}{c|}{33.1}                      & \multicolumn{1}{c|}{\textbf{38.4}}                 \\ \cline{2-4} 
\end{tabular}\end{center}}
\caption{\small{ The table shows the quantitative evaluation (DISTS \cite{ding2020image} and NIMA \cite{talebi2018nima} scores) and qualitative evaluation (User Study). It can be observed that Sem-CS (ours) outperforms baseline methods under comparison. }}
\label{table:nima_dist}
\end{table}
\begin{table}[!ht]\small{
\begin{center}\renewcommand{\arraystretch}{1.25}
\begin{tabular}{ccc}
                                         & \cellcolor[HTML]{EFEFEF}Gen-Art & \cellcolor[HTML]{FFFFC7}Sem-CS (ours) \\ \cline{2-3} 
\multicolumn{1}{l|}{\textit{DISTS}}      & \multicolumn{1}{c|}{0.20}                      & \multicolumn{1}{c|}{\textbf{0.33}}                 \\ \cline{2-3} 
\multicolumn{1}{l|}{\textit{NIMA}}       & \multicolumn{1}{c|}{4.34}                      & \multicolumn{1}{c|}{\textbf{5.52}}                 \\ \cline{2-3} 
\multicolumn{1}{l|}{\textit{User Study}} & \multicolumn{1}{c|}{48.2}                      & \multicolumn{1}{c|}{\textbf{51.7}}                 \\ \cline{2-3} 
\end{tabular}\end{center}}
\vspace{-0.4cm}
\caption{\small{\textbf{Ablation Study.} The figure shows the scores for ablation study performed for controllable generation using multiple text conditions}}
\label{table:ablation}
\end{table}
\vspace{-0.6cm}
\section{Conclusion}
We have proposed Semantic CLIPStyler (Sem-CS) to preserve the semantics of objects when performing text-guided style supervision in image style transfer. Quantitative and qualitative results show that Sem-CS achieves superior image style transfer and minimizes content mismatch in style transfer output. We also showed that the degree of style transfer can be controlled by incorporating global background and foreground loss. Different style texts can be applied to salient objects and background of the content image. This prevents the network from over-stylizing the content image and preserves the semantics of objects during style transfer. Future work would include extending this to handle more than two objects in the content image.

\bibliographystyle{IEEEbib}

\bibliography{refs}

\vfill\pagebreak
\begin{center}
    \Huge{\textbf{Appendix}}
\end{center}
\appendix
  \section{Salient Object Detection}
    \label{sec:sec_dssn}
    In this section, we describe phase 1 of Salient object detection.
   Let $I \in R^{3 \times M \times N}$ be content image (input).
    We extract deep patch features $f = \phi(I)$ of the input image I from the attention block of the last layer of $\phi$, a self-supervised transformer DINO-ViT-Base.
    Then, a weighted graph over patch features is built with the affinity between patch features as edge weights. 
    An affinity matrix is a weighted sum of semantic patch-wise features and colour matrices.
    \begin{equation}
        W=W_{feat} + \lambda_{knn}W_{knn}
    \end{equation}
    where $\lambda_{knn}$ is a parameter that regularizes semantic features and color consistency.
    Patch-wise features consists of only the correlated features, therefore $W_{feat}$ is thresholded at 0 as follows:
        \begin{equation}
        W_{feat}=ff^T \odot (ff^T > 0)
     \end{equation}
    Color affinity matrix between patch features u and v is calculated using sparse-KNN matting  as below. HSV color space is obtained using KNN-matting of the content image.
    \begin{equation}
        W_{knn}(u,v)=\begin{cases}
                                    1- \psi(u)-\psi(v), u \in KNN_{\psi}(v), \\
                                    0, otherwise
                                \end{cases}
    \end{equation}
    where, $\psi(u)$ is vector in $\mathcal{R}^5$ consisting of color and spatial information.
     Given W, the eigen decomposition of its graph laplacian L is considered for obtaining soft segments: $\{y_0, y_1· · ·, y_{n−1}\}$. We examine only Fiedler eigenvector $y_1$ that corresponds to semantic region and it usually identifies the most salient object in the image.

    \section{Other Loss}
    The content loss preserves the content during the transfer, and total variation regularization loss alleviates side artefacts from irregular pixels. Thus, total loss is calculated as follows:
       \begin{equation}
       \mathcal{L}_{total}=\lambda_{B}* \mathcal{L}_{BGlob} + \lambda_{F}*\mathcal{L}_{FGlob} + L_{others}
   \end{equation}
   where, 
  \begin{equation}
  \begin{split}
      \mathcal{L}_{total}=\ \lambda_C*L_{content} + \lambda_{TV}*L_{TV}
  \end{split}     
  \end{equation}

   where, $\lambda_C$ and $\lambda_{TV}$ are controlling parameters for content loss and TC regularization loss. 

    \section{Data Collection}
    To transfer artistic style from texts to content images, we tested 58 images: 40 of them are sceneries, and 18 of them are portraits. We had collected 50 famous text conditions. We then assigned them into eight different categories: abstract art, paintings,  portraits, sceneries, buildings and vehicles. We applied 50 style texts for each content image. Hence, we obtained 2900 photos in our final data set for single text condition. For double text conditions, we considered 10 content images and applied 2500 text conditions with all permutations and a combination of 50 text conditions and hence obtained 25000 images. 

\section{Implementation Details:}
Since CLIP model receives the images with resolution of 224×224, we resized
all of the images including patch and whole image before feeding to the CLIP model. For augmentations, we use perspective augmentation which is implemented on Pytorch torchvision library. We directly used RandomPerspective(distortion\_scale=0.5). The random perspective function is implemented in torchvision.transforms. We used batch size of 4 and Adam optimizer with learning rate of 1 × 10−4. Similar to our basic method, we used learning rate decay strategy. For augmentation, we applied random perspective augmentation for 16 times, therefore our total training patch number per iteration is N = 64. 

\section{Experimental Results}
In this section, we provide an extended version of the experiments performed in the paper. 
\begin{figure*}[!h]
   \centering
    \begin{minipage}{0.10\linewidth}
     \centering
        \textbf{Text Condition}
    \end{minipage}
    \hfill
    \begin{minipage}{0.21\linewidth}
     \centering
        \textbf{Input Image}
    \end{minipage}
    \hfill
    \begin{minipage}{0.21\linewidth}
     \centering
        CLIPStyler
    \end{minipage}
    \hfill
    \begin{minipage}{0.21\linewidth}
     \centering
        Generative Artisan
    \end{minipage}
    \hfill
    \begin{minipage}{0.21\linewidth}
     \centering
        Sem-CS (ours)
    \end{minipage}

            \fbox{
        \begin{minipage}[c][1.8cm][c]{1.8cm}
         \centering
          A monet style painting
        \end{minipage}}
        \begin{minipage}{0.21\linewidth}
         \centering
          \includegraphics[width=0.99\linewidth]{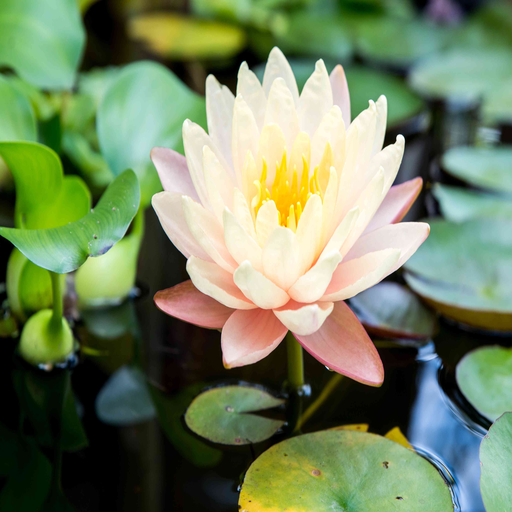}
        \end{minipage}
        \hfill
        \begin{minipage}{0.21\linewidth}
         \centering
          \includegraphics[width=0.99\linewidth]{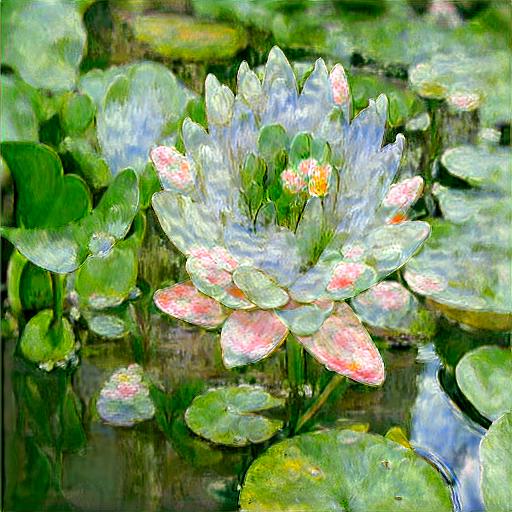}
        \end{minipage}
        \begin{minipage}{0.21\linewidth}
         \centering
          \includegraphics[width=0.99\linewidth]{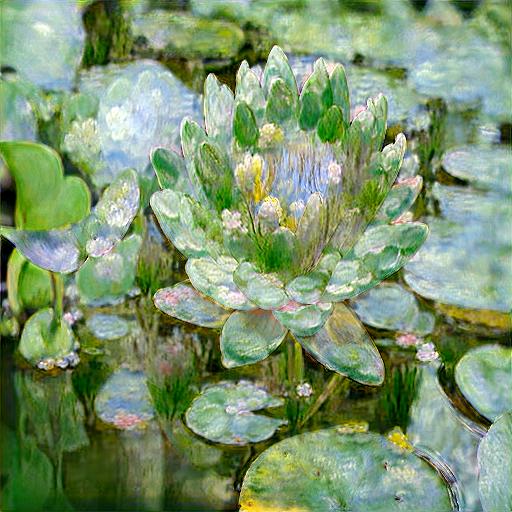}
        \end{minipage}
        \begin{minipage}{0.21\linewidth}
         \centering
          \includegraphics[width=0.99\linewidth]{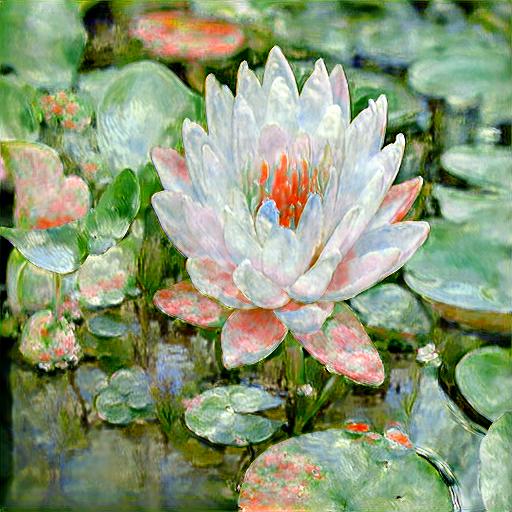}
        \end{minipage}
        \fbox{
        \begin{minipage}[c][1.8cm][c]{1.8cm}
         \centering
         \normalsize
            A graffiti style painting
        \end{minipage}}
        \begin{minipage}{0.21\textwidth}
         \centering             \includegraphics[width=0.99\linewidth]{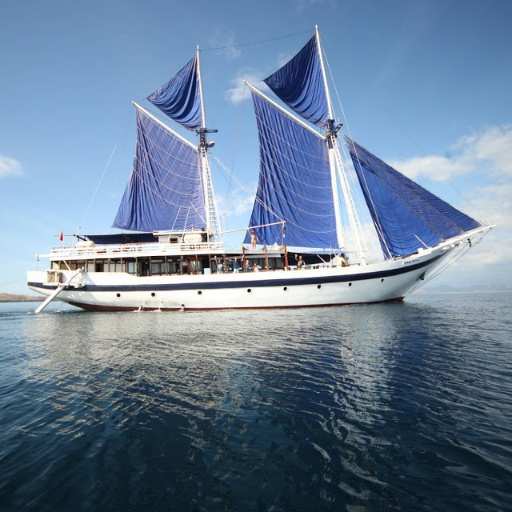}
        \end{minipage}
        \hfill
        \begin{minipage}{0.21\textwidth}
         \centering
             \includegraphics[width=0.99\linewidth]{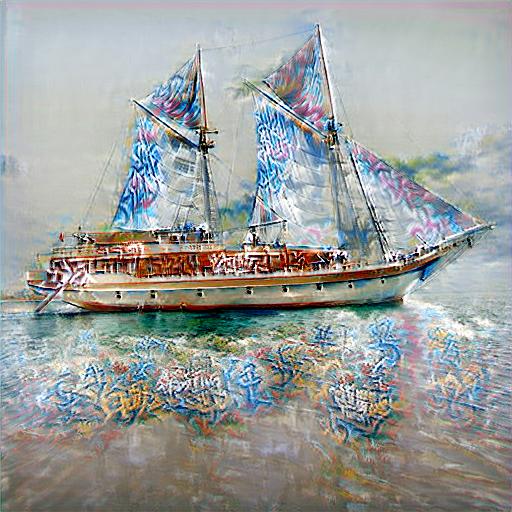}
        \end{minipage}
        \begin{minipage}{0.21\textwidth}
         \centering               
        \includegraphics[width=0.99\linewidth]{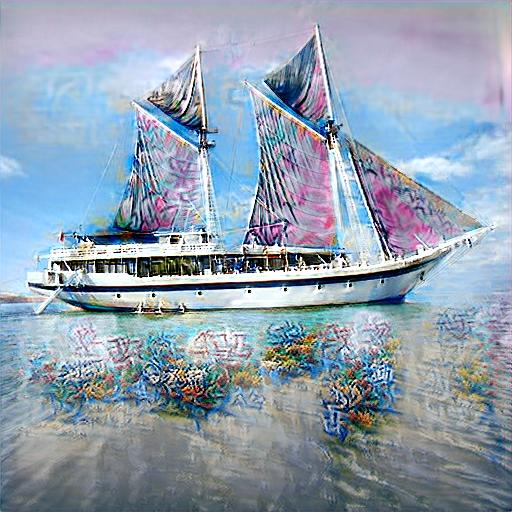}
        \end{minipage}
        \begin{minipage}{0.21\textwidth}
         \centering                 
            \includegraphics[width=0.99\linewidth]{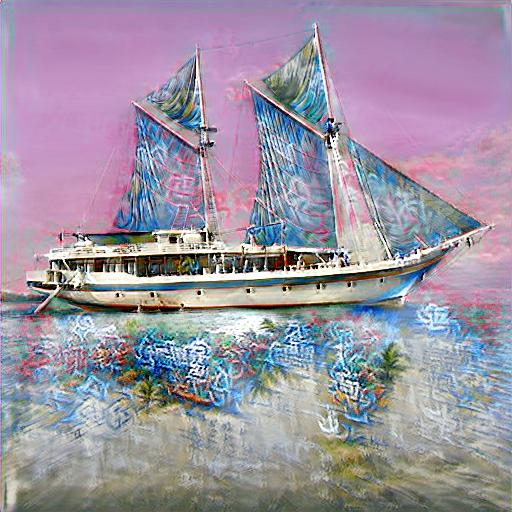}
        \end{minipage}
        \fbox{
         \begin{minipage}[c][1.8cm][c]{1.8cm}
         \centering
         \normalsize
            Acrylic painting
        \end{minipage}}
        \begin{minipage}{0.21\textwidth}
         \centering
             \includegraphics[width=0.99\linewidth]{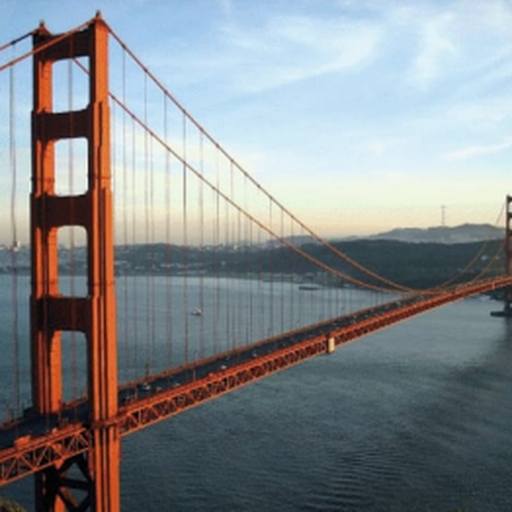}
        \end{minipage}
        \hfill
        \begin{minipage}{0.21\textwidth}
         \centering
             \includegraphics[width=0.99\linewidth]{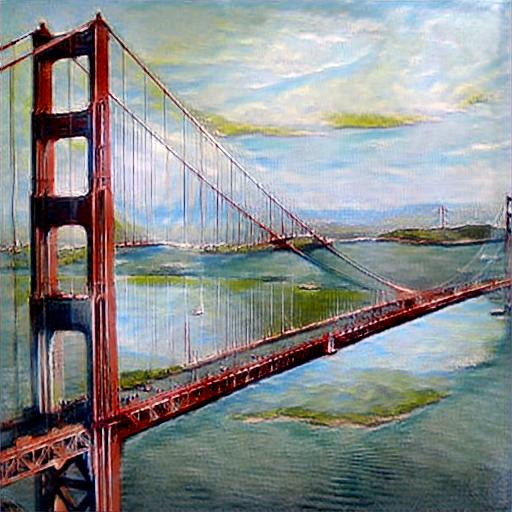}
        \end{minipage}
        \begin{minipage}{0.21\textwidth}
         \centering                 
            \includegraphics[width=0.99\linewidth]{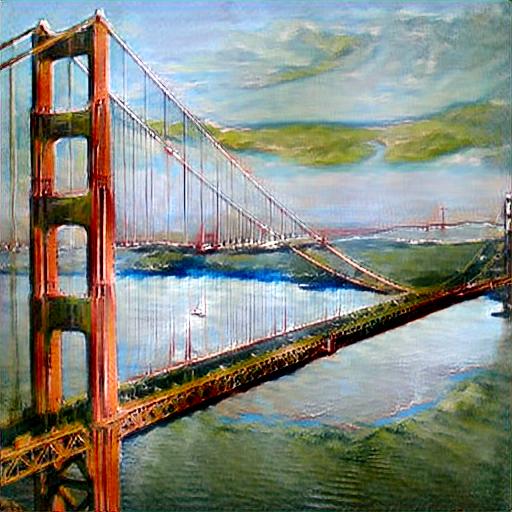}
        \end{minipage}
        \begin{minipage}{0.21\textwidth}
         \centering                 
            \includegraphics[width=0.99\linewidth]{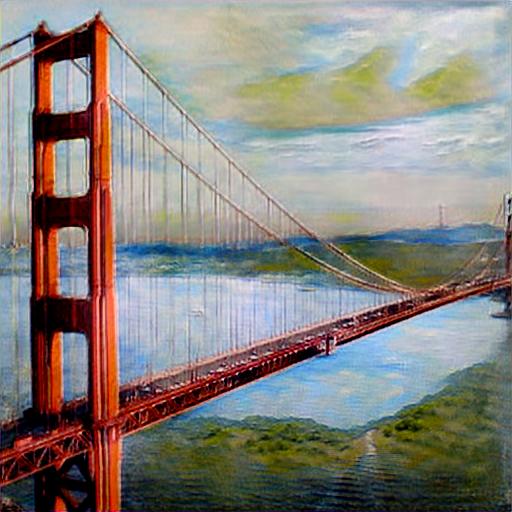}
        \end{minipage}
     \fbox{
         \begin{minipage}[c][1.8cm][c]{1.8cm}
         \centering
         \normalsize
            Snowy
        \end{minipage}}
        \begin{minipage}{0.21\textwidth}
         \centering
             \includegraphics[width=0.99\linewidth]{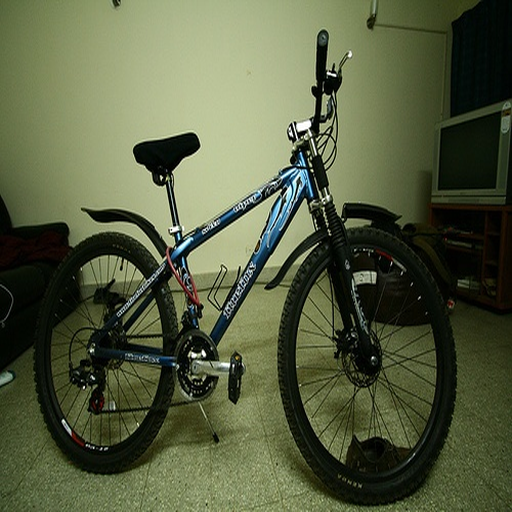}
        \end{minipage}
        \hfill
        \begin{minipage}{0.21\textwidth}
         \centering
             \includegraphics[width=0.99\linewidth]{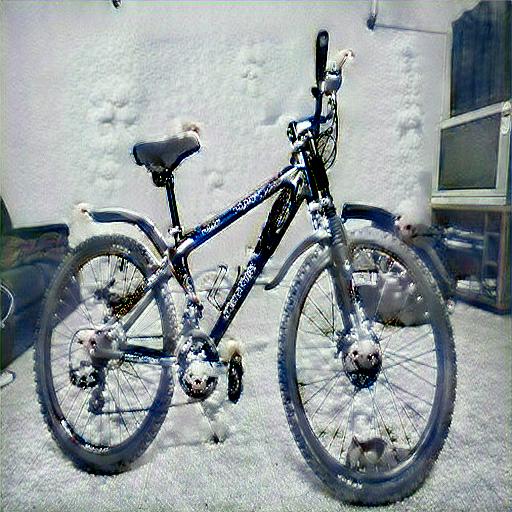}
        \end{minipage}
        \begin{minipage}{0.21\textwidth}
         \centering                 
            \includegraphics[width=0.99\linewidth]{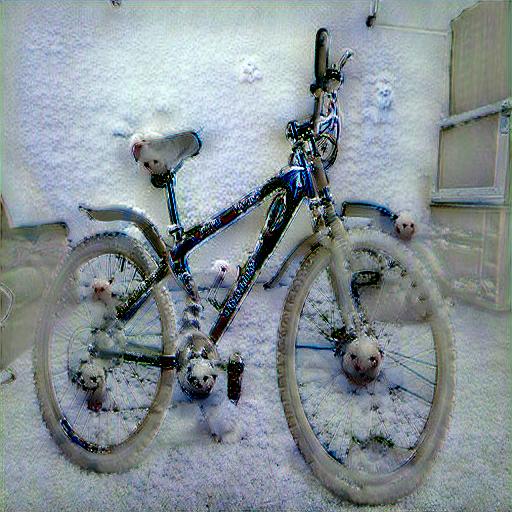}
        \end{minipage}
        \begin{minipage}{0.21\textwidth}
         \centering                 
            \includegraphics[width=0.99\linewidth]{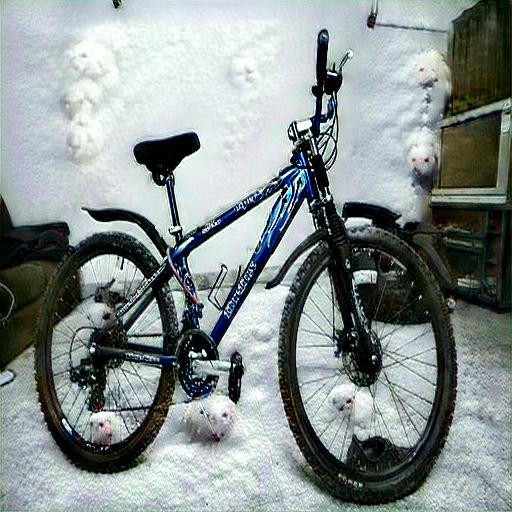}
        \end{minipage}
                \fbox{
         \begin{minipage}[c][1.8cm][c]{1.8cm}
         \centering
         \normalsize
            Desert Sand
        \end{minipage}}
        \begin{minipage}{0.21\textwidth}
         \centering
             \includegraphics[width=0.99\linewidth]{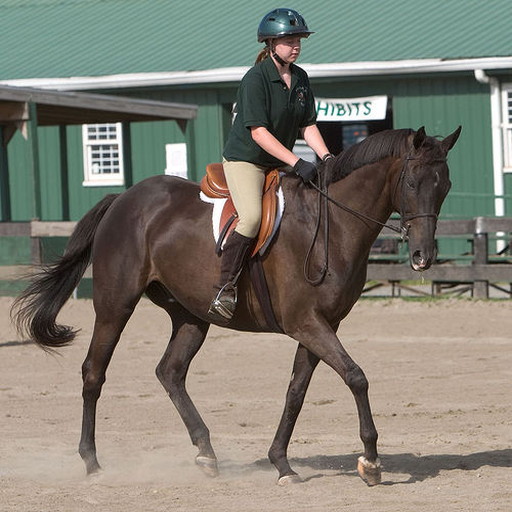}
        \end{minipage}
        \hfill
        \begin{minipage}{0.21\textwidth}
         \centering
             \includegraphics[width=0.99\linewidth]{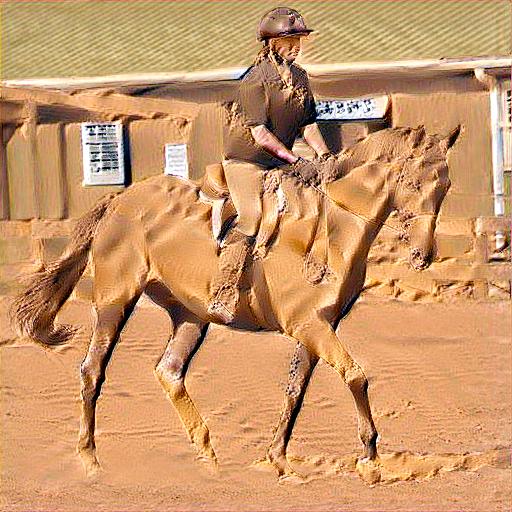}
        \end{minipage}
        \begin{minipage}{0.21\textwidth}
         \centering                 
            \includegraphics[width=0.99\linewidth]{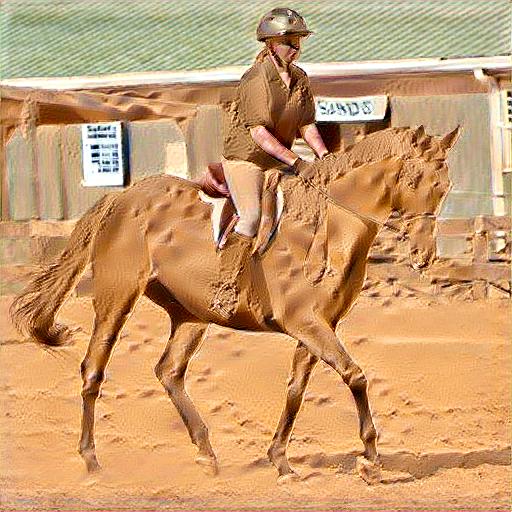}
        \end{minipage}
        \begin{minipage}{0.21\textwidth}
         \centering                 
            \includegraphics[width=0.99\linewidth]{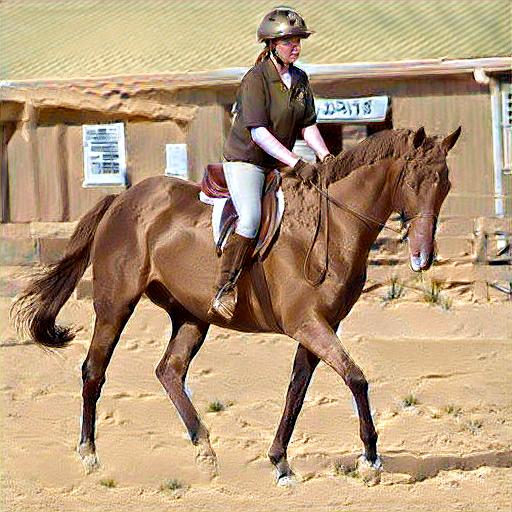}
        \end{minipage}
        \fbox{
             \begin{minipage}[c][1.8cm][c]{1.8cm}
         \centering
         \normalsize
            A fauvism style painting
        \end{minipage}}
        \begin{minipage}{0.21\textwidth}
         \centering
             \includegraphics[width=0.99\linewidth]{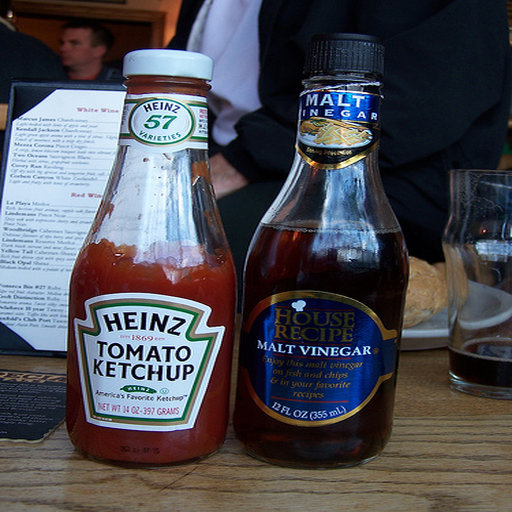}
        \end{minipage}
        \hfill
        \begin{minipage}{0.21\textwidth}
         \centering
             \includegraphics[width=0.99\linewidth]{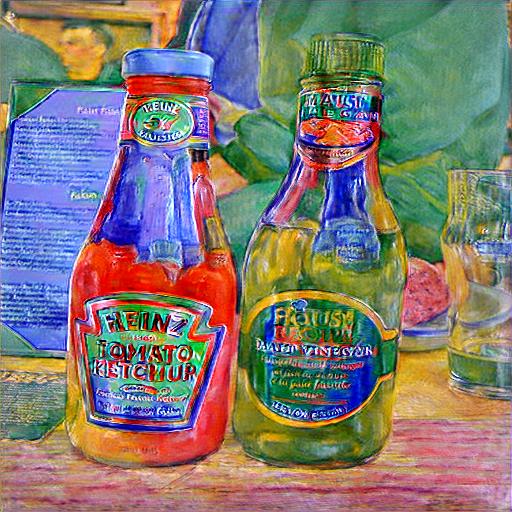}
        \end{minipage}
        \begin{minipage}{0.21\textwidth}
         \centering                 
            \includegraphics[width=0.99\linewidth]{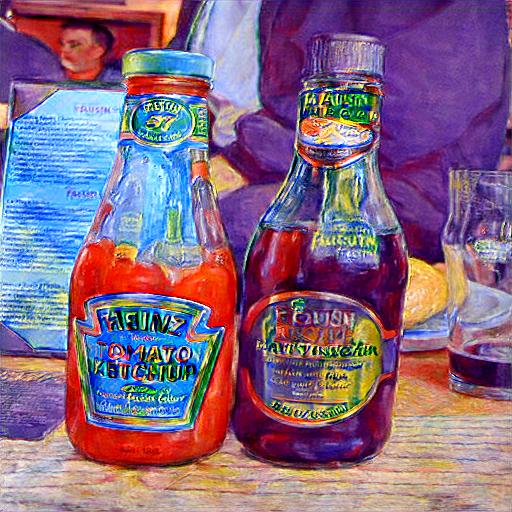}
        \end{minipage}
        \begin{minipage}{0.21\textwidth}
         \centering                 
            \includegraphics[width=0.99\linewidth]{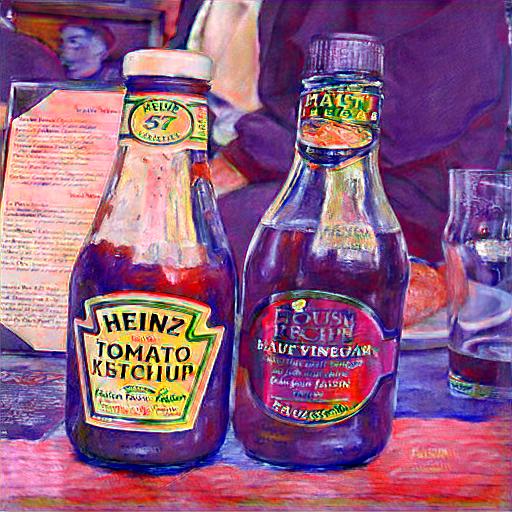}
        \end{minipage}  
    \end{figure*}

    \begin{figure*}[!h]
       \centering
        \begin{minipage}{0.10\linewidth}
         \centering
            \textbf{Text Condition}
        \end{minipage}
        \hfill
        \begin{minipage}{0.21\linewidth}
         \centering
            \textbf{Input Image}
        \end{minipage}
        \hfill
        \begin{minipage}{0.21\linewidth}
         \centering
            CLIPStyler
        \end{minipage}
        \hfill
        \begin{minipage}{0.21\linewidth}
         \centering
            Generative Artisan
        \end{minipage}
        \hfill
        \begin{minipage}{0.21\linewidth}
         \centering
            Sem-CS (ours)
        \end{minipage}        
        \fbox{
             \begin{minipage}[c][1.8cm][c]{1.8cm}
         \centering
         \normalsize
            Red rocks
        \end{minipage}}
        \begin{minipage}{0.21\textwidth}
         \centering
             \includegraphics[width=0.99\linewidth]{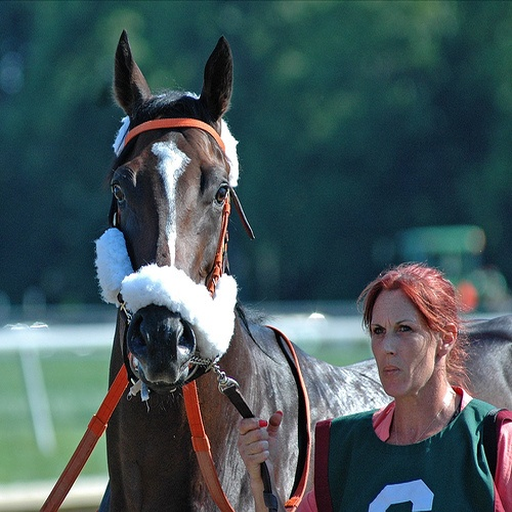}
        \end{minipage}
        \hfill
        \begin{minipage}{0.21\textwidth}
         \centering
             \includegraphics[width=0.99\linewidth]{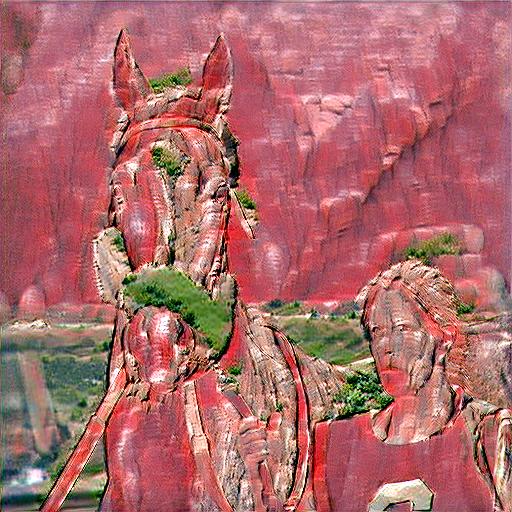}
        \end{minipage}
        \begin{minipage}{0.21\textwidth}
         \centering                 
            \includegraphics[width=0.99\linewidth]{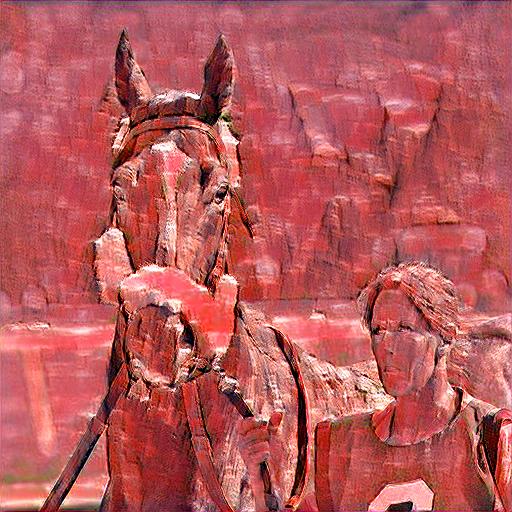}
        \end{minipage}
        \begin{minipage}{0.21\textwidth}
         \centering                 
            \includegraphics[width=0.99\linewidth]{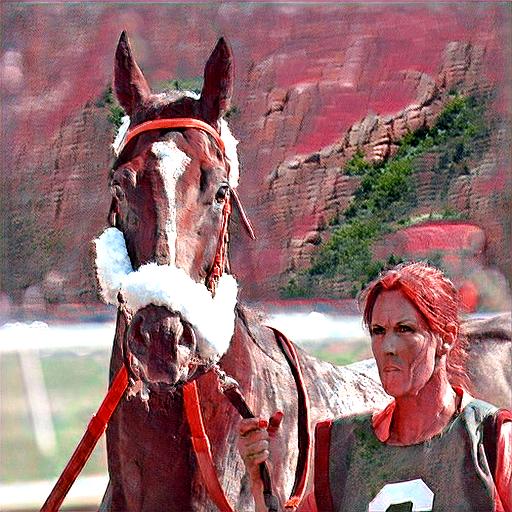}
        \end{minipage}
      \fbox{
             \begin{minipage}[c][1.8cm][c]{1.8cm}
         \centering
         \small
         \normalsize
            A watercolor painting with purple brush
        \end{minipage}}
        \begin{minipage}{0.21\textwidth}
         \centering
             \includegraphics[width=0.99\linewidth]{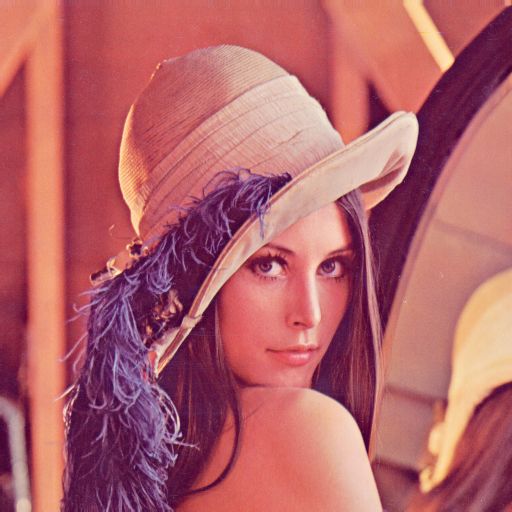}
        \end{minipage}
        \hfill
        \begin{minipage}{0.21\textwidth}
         \centering
             \includegraphics[width=0.99\linewidth]{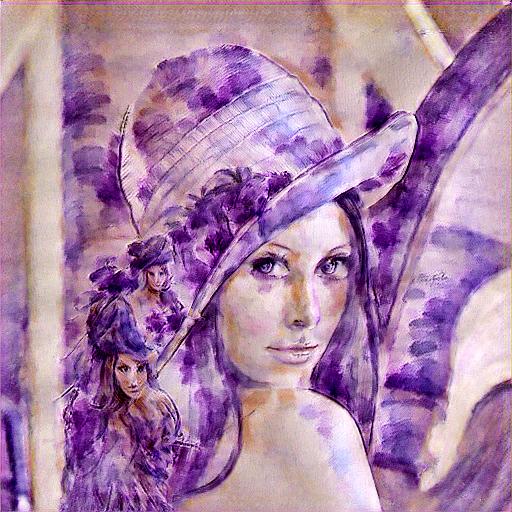}
        \end{minipage}
        \begin{minipage}{0.21\textwidth}
         \centering                 
            \includegraphics[width=0.99\linewidth]{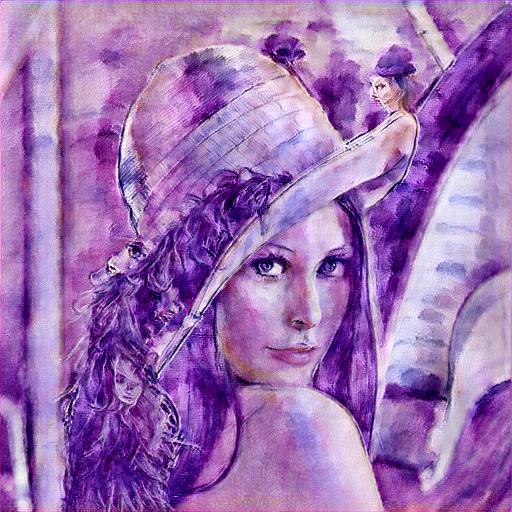}
        \end{minipage}
        \begin{minipage}{0.21\textwidth}
         \centering                 
            \includegraphics[width=0.99\linewidth]{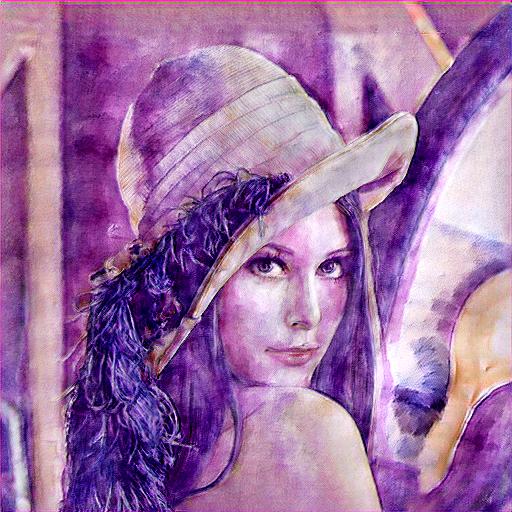}
        \end{minipage}
       \fbox{
             \begin{minipage}[c][1.8cm][c]{1.8cm}
         \centering
         \normalsize
            An oil painting of white roses
        \end{minipage}}
        \begin{minipage}{0.21\textwidth}
         \centering
             \includegraphics[width=0.99\linewidth]{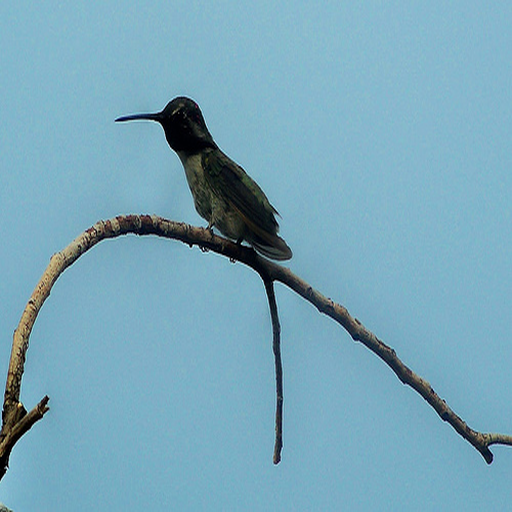}
        \end{minipage}
        \hfill
        \begin{minipage}{0.21\textwidth}
         \centering
             \includegraphics[width=0.99\linewidth]{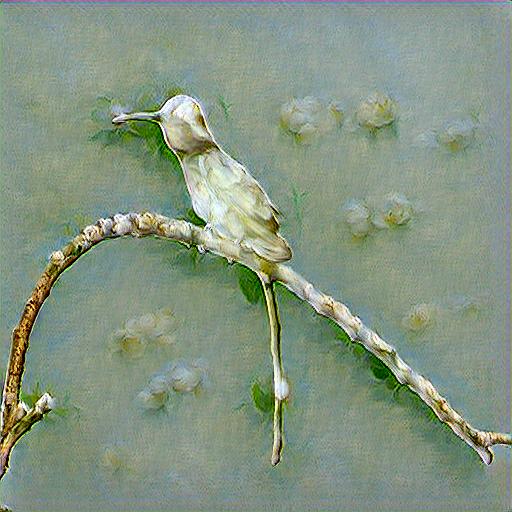}
        \end{minipage}
        \begin{minipage}{0.21\textwidth}
         \centering                 
            \includegraphics[width=0.99\linewidth]{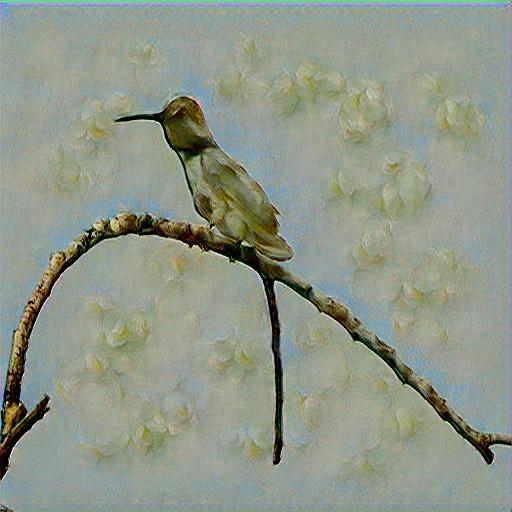}
        \end{minipage}
        \begin{minipage}{0.21\textwidth}
         \centering                 
            \includegraphics[width=0.99\linewidth]{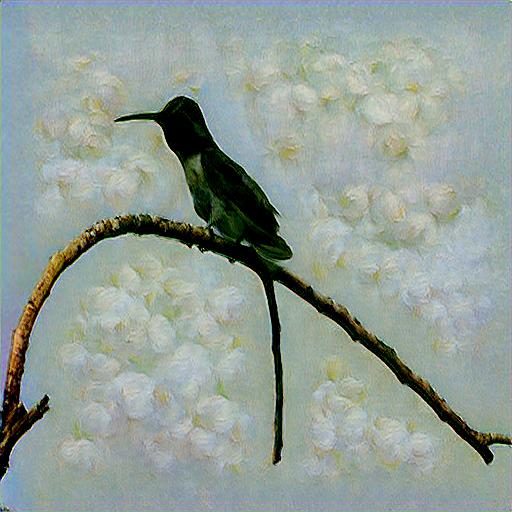}
        \end{minipage}
      \fbox{
             \begin{minipage}[c][1.8cm][c]{1.8cm}
         \centering
         \small
         \normalsize
            A watercolor painting of leaf
        \end{minipage}}
        \begin{minipage}{0.21\textwidth}
         \centering
             \includegraphics[width=0.99\linewidth]{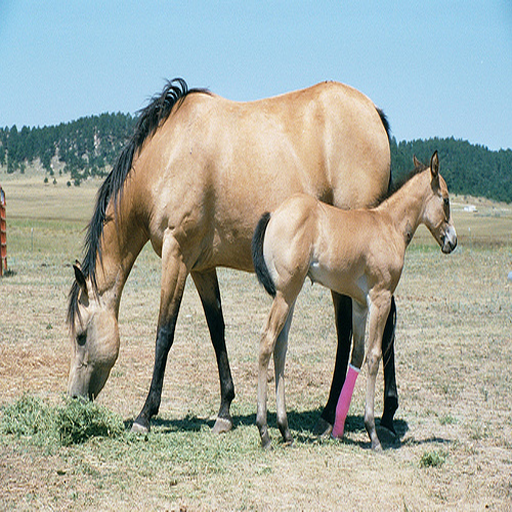}
        \end{minipage}
        \hfill
        \begin{minipage}{0.21\textwidth}
         \centering
             \includegraphics[width=0.99\linewidth]{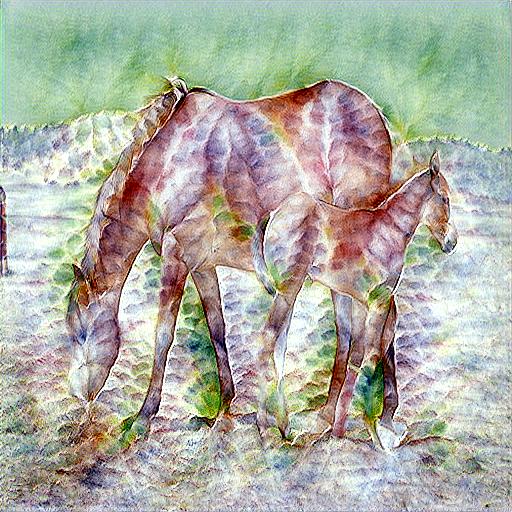}
        \end{minipage}
        \begin{minipage}{0.21\textwidth}
         \centering                 
            \includegraphics[width=0.99\linewidth]{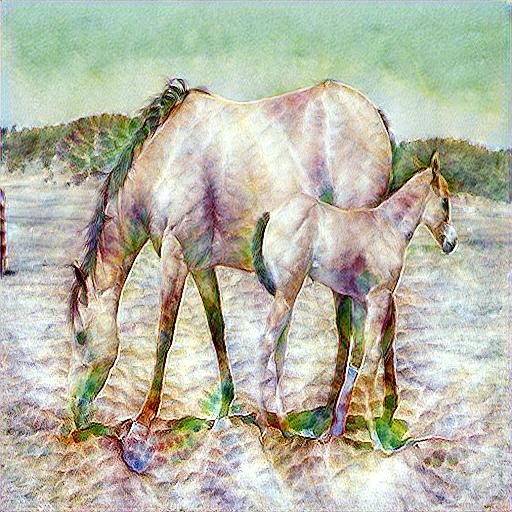}
        \end{minipage}
        \begin{minipage}{0.21\textwidth}
         \centering                 
            \includegraphics[width=0.99\linewidth]{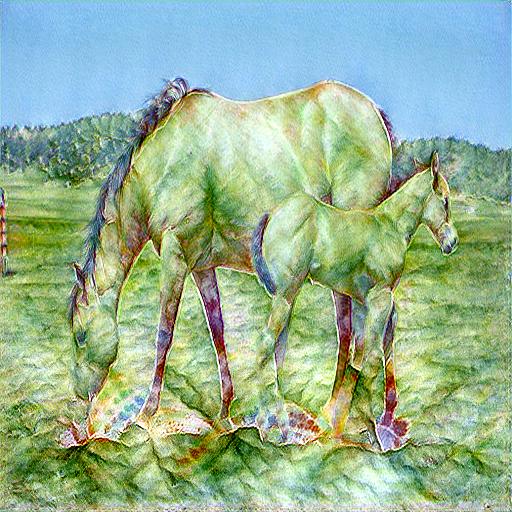}
        \end{minipage}
        
\end{figure*}   
\vspace{0.2cm}     
   \hrule
   \vspace{0.2cm}
\begin{figure*}[!htb]
        \begin{minipage}{0.10\linewidth}
         \centering
         \small
            \textbf{Text Condition}
        \end{minipage}
        \begin{minipage}{0.27\linewidth}
         \centering
         \small
            \textbf{Input Image}
        \end{minipage}
      \begin{minipage}{0.27\linewidth}
         \centering
         \small
            Generative Artisan
        \end{minipage}
      \begin{minipage}{0.27\linewidth}
         \centering
         \small
            Sem-CS (ours)
        \end{minipage}    
     \fbox{
             \begin{minipage}[c][1.8cm][c]{1.8cm}
         \centering
         \footnotesize
            \textbf{F:}  Pop Art\\~
            \textbf{B:} Starry Night by Vincent Van Gogh
        \end{minipage}}
        \begin{minipage}{0.27\linewidth}
         \centering
             \includegraphics[width=0.99\linewidth]{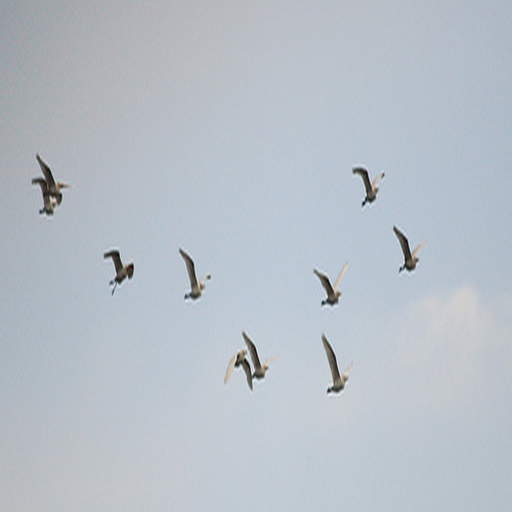}
        \end{minipage}
        \hfill
        \begin{minipage}{0.27\linewidth}
         \centering
             \includegraphics[width=0.99\linewidth]{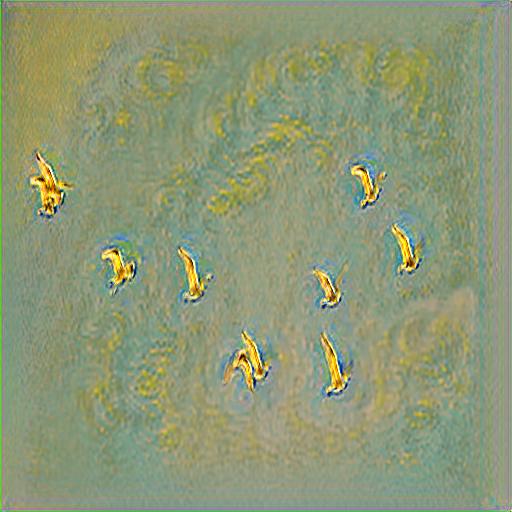}
        \end{minipage}
        \begin{minipage}{0.27\linewidth}
         \centering
             \includegraphics[width=0.99\linewidth]{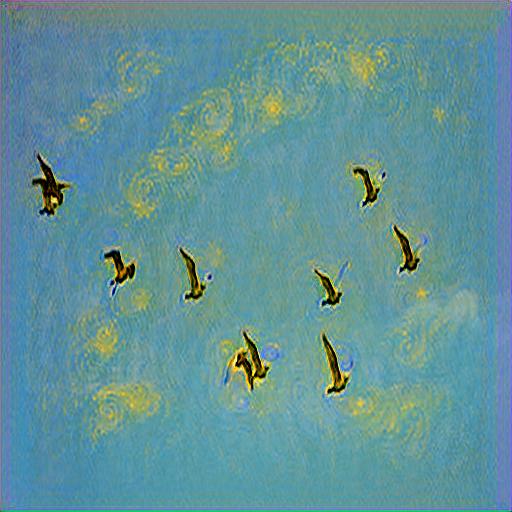}
        \end{minipage}
       \fbox{
             \begin{minipage}[c][1.8cm][c]{1.8cm}
         \centering
         \footnotesize
            \textbf{F:} Red Rocks\\~
            \textbf{B:} Snowy
        \end{minipage}}
        \begin{minipage}{0.27\linewidth}
         \centering
             \includegraphics[width=0.99\linewidth]{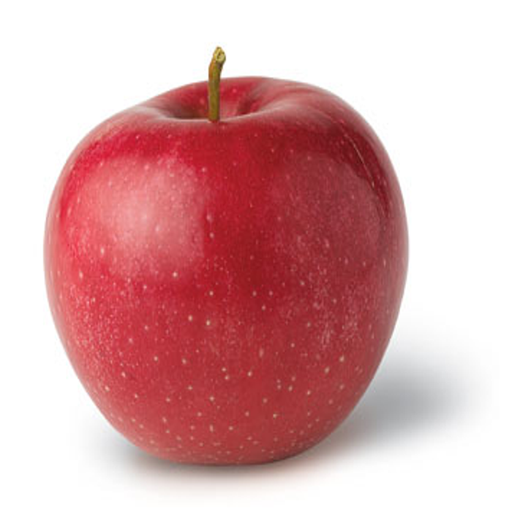}
        \end{minipage}
        \hfill
        \begin{minipage}{0.27\linewidth}
         \centering
             \includegraphics[width=0.99\linewidth]{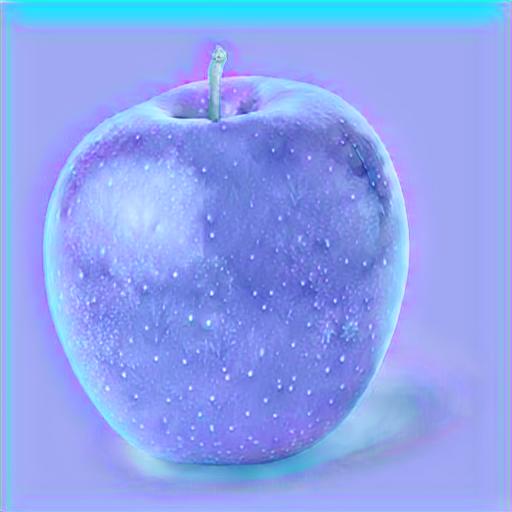}
        \end{minipage}
        \begin{minipage}{0.27\linewidth}
         \centering
             \includegraphics[width=0.99\linewidth]{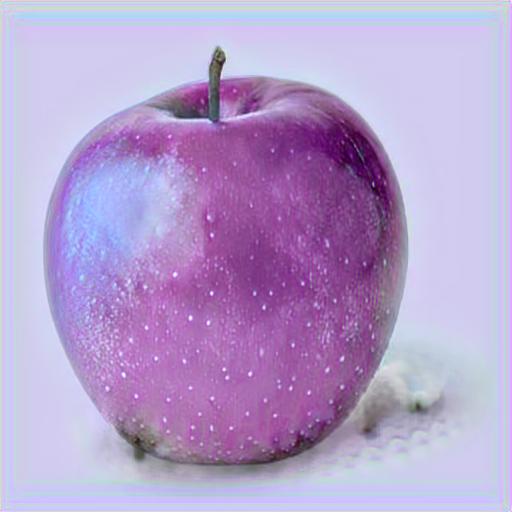}
        \end{minipage}
        
    \end{figure*}

\begin{table}[!ht]
\caption{DISTS}
\begin{tabular}{|lll|ll|}
\hline
\multicolumn{3}{|c|}{\textbf{One Style Text}}                                                                            & \multicolumn{2}{c|}{\textbf{Two Style Texts}}                                \\ \hline
\multicolumn{1}{|c|}{\textbf{Sem-CS}} & \multicolumn{1}{c|}{\textbf{Gen-Art}} & \multicolumn{1}{c|}{\textbf{CLIPStyler}} & \multicolumn{1}{c|}{\textbf{Sem-CS}} & \multicolumn{1}{c|}{\textbf{Gen-Art}} \\ \hline
\multicolumn{1}{|l|}{0.39}            & \multicolumn{1}{l|}{0.22}             & 0.36                                     & \multicolumn{1}{l|}{0.43}            & 0.19                                  \\ \hline
\multicolumn{1}{|l|}{0.38}            & \multicolumn{1}{l|}{0.22}             & 0.25                                     & \multicolumn{1}{l|}{0.41}            & 0.22                                  \\ \hline
\multicolumn{1}{|l|}{0.33}            & \multicolumn{1}{l|}{0.19}             & 0.37                                     & \multicolumn{1}{l|}{0.45}            & 0.27                                  \\ \hline
\multicolumn{1}{|l|}{0.31}            & \multicolumn{1}{l|}{0.18}             & 0.31                                     & \multicolumn{1}{l|}{0.32}            & 0.14                                  \\ \hline
\multicolumn{1}{|l|}{0.34}            & \multicolumn{1}{l|}{0.21}             & 0.29                                     & \multicolumn{1}{l|}{0.41}            & 0.23                                  \\ \hline
\multicolumn{1}{|l|}{0.34}            & \multicolumn{1}{l|}{0.21}             & 0.35                                     & \multicolumn{1}{l|}{0.33}            & 0.16                                  \\ \hline
\multicolumn{1}{|l|}{0.31}            & \multicolumn{1}{l|}{0.19}             & 0.18                                     & \multicolumn{1}{l|}{0.45}            & 0.28                                  \\ \hline
\multicolumn{1}{|l|}{0.38}            & \multicolumn{1}{l|}{0.26}             & 0.33                                     & \multicolumn{1}{l|}{0.42}            & 0.25                                  \\ \hline
\multicolumn{1}{|l|}{0.38}            & \multicolumn{1}{l|}{0.26}             & 0.37                                     & \multicolumn{1}{l|}{0.28}            & 0.12                                  \\ \hline
\multicolumn{1}{|l|}{0.35}            & \multicolumn{1}{l|}{0.23}             & 0.35                                     & \multicolumn{1}{l|}{0.31}            & 0.15                                  \\ \hline
\multicolumn{1}{|l|}{0.32}            & \multicolumn{1}{l|}{0.22}             & 0.30                                     & \multicolumn{1}{l|}{0.31}            & 0.15                                  \\ \hline
\multicolumn{1}{|l|}{0.35}            & \multicolumn{1}{l|}{0.24}             & 0.38                                     & \multicolumn{1}{l|}{0.44}            & 0.28                                  \\ \hline
\multicolumn{1}{|l|}{0.33}            & \multicolumn{1}{l|}{0.22}             & 0.31                                     & \multicolumn{1}{l|}{0.44}            & 0.28                                  \\ \hline
\multicolumn{1}{|l|}{0.31}            & \multicolumn{1}{l|}{0.21}             & 0.28                                     & \multicolumn{1}{l|}{0.29}            & 0.13                                  \\ \hline
\multicolumn{1}{|l|}{0.40}            & \multicolumn{1}{l|}{0.30}             & 0.38                                     & \multicolumn{1}{l|}{0.41}            & 0.25                                  \\ \hline
\multicolumn{1}{|l|}{0.38}            & \multicolumn{1}{l|}{0.28}             & 0.39                                     & \multicolumn{1}{l|}{0.28}            & 0.13                                  \\ \hline
\multicolumn{1}{|l|}{0.40}            & \multicolumn{1}{l|}{0.30}             & 0.37                                     & \multicolumn{1}{l|}{0.28}            & 0.13                                  \\ \hline
\multicolumn{1}{|l|}{0.41}            & \multicolumn{1}{l|}{0.31}             & 0.41                                     & \multicolumn{1}{l|}{0.31}            & 0.16                                  \\ \hline
\multicolumn{1}{|l|}{0.27}            & \multicolumn{1}{l|}{0.18}             & 0.33                                     & \multicolumn{1}{l|}{0.43}            & 0.28                                  \\ \hline
\multicolumn{1}{|l|}{0.31}            & \multicolumn{1}{l|}{0.22}             & 0.32                                     & \multicolumn{1}{l|}{0.45}            & 0.31                                  \\ \hline
\multicolumn{1}{|l|}{0.36}            & \multicolumn{1}{l|}{0.26}             & 0.40                                     & \multicolumn{1}{l|}{0.39}            & 0.25                                  \\ \hline
\multicolumn{1}{|l|}{0.36}            & \multicolumn{1}{l|}{0.27}             & 0.42                                     & \multicolumn{1}{l|}{0.39}            & 0.25                                  \\ \hline
\multicolumn{1}{|l|}{0.31}            & \multicolumn{1}{l|}{0.22}             & 0.27                                     & \multicolumn{1}{l|}{0.27}            & 0.13                                  \\ \hline
\multicolumn{1}{|l|}{0.30}            & \multicolumn{1}{l|}{0.21}             & 0.22                                     & \multicolumn{1}{l|}{0.33}            & 0.19                                  \\ \hline
\multicolumn{1}{|l|}{0.36}            & \multicolumn{1}{l|}{0.27}             & 0.39                                     & \multicolumn{1}{l|}{0.4}             & 0.26                                  \\ \hline
\multicolumn{1}{|l|}{0.37}            & \multicolumn{1}{l|}{0.29}             & 0.30                                     & \multicolumn{1}{l|}{0.36}            & 0.22                                  \\ \hline
\multicolumn{1}{|l|}{0.30}            & \multicolumn{1}{l|}{0.21}             & 0.36                                     & \multicolumn{1}{l|}{0.35}            & 0.21                                  \\ \hline
\multicolumn{1}{|l|}{0.35}            & \multicolumn{1}{l|}{0.26}             & 0.35                                     & \multicolumn{1}{l|}{0.42}            & 0.28                                  \\ \hline
\multicolumn{1}{|l|}{0.43}            & \multicolumn{1}{l|}{0.34}             & 0.40                                     & \multicolumn{1}{l|}{0.28}            & 0.15                                  \\ \hline
\multicolumn{1}{|l|}{0.29}            & \multicolumn{1}{l|}{0.21}             & 0.26                                     & \multicolumn{1}{l|}{0.28}            & 0.15                                  \\ \hline
\multicolumn{1}{|l|}{0.37}            & \multicolumn{1}{l|}{0.28}             & 0.35                                     & \multicolumn{1}{l|}{0.28}            & 0.15                                  \\ \hline
\multicolumn{1}{|l|}{0.29}            & \multicolumn{1}{l|}{0.21}             & 0.23                                     & \multicolumn{1}{l|}{0.28}            & 0.15                                  \\ \hline
\multicolumn{1}{|l|}{0.32}            & \multicolumn{1}{l|}{0.23}             & 0.28                                     & \multicolumn{1}{l|}{0.27}            & 0.14                                  \\ \hline
\multicolumn{1}{|l|}{0.42}            & \multicolumn{1}{l|}{0.34}             & 0.40                                     & \multicolumn{1}{l|}{0.27}            & 0.14                                  \\ \hline
\multicolumn{1}{|l|}{0.32}            & \multicolumn{1}{l|}{0.24}             & 0.27                                     & \multicolumn{1}{l|}{0.44}            & 0.31                                  \\ \hline
\multicolumn{1}{|l|}{0.27}            & \multicolumn{1}{l|}{0.19}             & 0.32                                     & \multicolumn{1}{l|}{0.44}            & 0.31                                  \\ \hline
\multicolumn{1}{|l|}{0.24}            & \multicolumn{1}{l|}{0.16}             & 0.31                                     & \multicolumn{1}{l|}{0.31}            & 0.18                                  \\ \hline
\multicolumn{1}{|l|}{0.35}            & \multicolumn{1}{l|}{0.26}             & 0.38                                     & \multicolumn{1}{l|}{0.27}            & 0.14                                  \\ \hline
\multicolumn{1}{|l|}{0.39}            & \multicolumn{1}{l|}{0.31}             & 0.37                                     & \multicolumn{1}{l|}{0.27}            & 0.14                                  \\ \hline
\multicolumn{1}{|l|}{0.46}            & \multicolumn{1}{l|}{0.38}             & 0.45                                     & \multicolumn{1}{l|}{0.25}            & 0.12                                  \\ \hline
\multicolumn{1}{|l|}{0.31}            & \multicolumn{1}{l|}{0.23}             & 0.35                                     & \multicolumn{1}{l|}{0.32}            & 0.19                                  \\ \hline
\multicolumn{1}{|l|}{0.36}            & \multicolumn{1}{l|}{0.28}             & 0.30                                     & \multicolumn{1}{l|}{0.27}            & 0.14                                  \\ \hline
\multicolumn{1}{|l|}{0.34}            & \multicolumn{1}{l|}{0.26}             & 0.25                                     & \multicolumn{1}{l|}{0.4}             & 0.27                                  \\ \hline
\multicolumn{1}{|l|}{0.28}            & \multicolumn{1}{l|}{0.20}             & 0.29                                     & \multicolumn{1}{l|}{0.38}            & 0.25                                  \\ \hline
\multicolumn{1}{|l|}{0.26}            & \multicolumn{1}{l|}{0.18}             & 0.33                                     & \multicolumn{1}{l|}{0.26}            & 0.13                                  \\ \hline
\multicolumn{1}{|l|}{0.36}            & \multicolumn{1}{l|}{0.28}             & 0.37                                     & \multicolumn{1}{l|}{0.29}            & 0.16                                  \\ \hline
\multicolumn{1}{|l|}{0.37}            & \multicolumn{1}{l|}{0.29}             & 0.34                                     & \multicolumn{1}{l|}{0.3}             & 0.17                                  \\ \hline
\multicolumn{1}{|l|}{0.31}            & \multicolumn{1}{l|}{0.23}             & 0.28                                     & \multicolumn{1}{l|}{0.36}            & 0.23                                  \\ \hline
\multicolumn{1}{|l|}{0.43}            & \multicolumn{1}{l|}{0.35}             & 0.43                                     & \multicolumn{1}{l|}{0.3}             & 0.17                                  \\ \hline
\multicolumn{1}{|l|}{0.29}            & \multicolumn{1}{l|}{0.21}             & 0.21                                     & \multicolumn{1}{l|}{0.41}            & 0.28                                  \\ \hline
\multicolumn{1}{|l|}{0.32}            & \multicolumn{1}{l|}{0.25}             & 0.30                                     & \multicolumn{1}{l|}{0.27}            & 0.15                                  \\ \hline
\end{tabular}
\end{table}

\begin{table}[!ht]
\begin{tabular}{|lll|ll|}
\hline
\multicolumn{3}{|c|}{\textbf{One Style Text}}                                                                            & \multicolumn{2}{c|}{\textbf{Two Style Texts}}                                \\ \hline
\multicolumn{1}{|c|}{\textbf{Sem-CS}} & \multicolumn{1}{c|}{\textbf{Gen-Art}} & \multicolumn{1}{c|}{\textbf{CLIPStyler}} & \multicolumn{1}{c|}{\textbf{Sem-CS}} & \multicolumn{1}{c|}{\textbf{Gen-Art}} \\ \hline
\multicolumn{1}{|l|}{0.28}            & \multicolumn{1}{l|}{0.20}             & 0.27                                     & \multicolumn{1}{l|}{0.27}            & 0.15                                  \\ \hline
\multicolumn{1}{|l|}{0.39}            & \multicolumn{1}{l|}{0.31}             & 0.28                                     & \multicolumn{1}{l|}{0.28}            & 0.16                                  \\ \hline
\multicolumn{1}{|l|}{0.32}            & \multicolumn{1}{l|}{0.25}             & 0.29                                     & \multicolumn{1}{l|}{0.27}            & 0.15                                  \\ \hline
\multicolumn{1}{|l|}{0.47}            & \multicolumn{1}{l|}{0.40}             & 0.44                                     & \multicolumn{1}{l|}{0.38}            & 0.26                                  \\ \hline
\multicolumn{1}{|l|}{0.27}            & \multicolumn{1}{l|}{0.19}             & 0.26                                     & \multicolumn{1}{l|}{0.32}            & 0.2                                   \\ \hline
\multicolumn{1}{|l|}{0.42}            & \multicolumn{1}{l|}{0.35}             & 0.33                                     & \multicolumn{1}{l|}{0.31}            & 0.19                                  \\ \hline
\multicolumn{1}{|l|}{0.25}            & \multicolumn{1}{l|}{0.18}             & 0.32                                     & \multicolumn{1}{l|}{0.4}             & 0.28                                  \\ \hline
\multicolumn{1}{|l|}{0.32}            & \multicolumn{1}{l|}{0.24}             & 0.40                                     & \multicolumn{1}{l|}{0.25}            & 0.13                                  \\ \hline
\multicolumn{1}{|l|}{0.40}            & \multicolumn{1}{l|}{0.33}             & 0.36                                     & \multicolumn{1}{l|}{0.38}            & 0.26                                  \\ \hline
\multicolumn{1}{|l|}{0.48}            & \multicolumn{1}{l|}{0.41}             & 0.51                                     & \multicolumn{1}{l|}{0.32}            & 0.2                                   \\ \hline
\multicolumn{1}{|l|}{0.35}            & \multicolumn{1}{l|}{0.28}             & 0.32                                     & \multicolumn{1}{l|}{0.49}            & 0.37                                  \\ \hline
\multicolumn{1}{|l|}{0.25}            & \multicolumn{1}{l|}{0.18}             & 0.21                                     & \multicolumn{1}{l|}{0.41}            & 0.29                                  \\ \hline
\multicolumn{1}{|l|}{0.39}            & \multicolumn{1}{l|}{0.32}             & 0.39                                     & \multicolumn{1}{l|}{0.4}             & 0.28                                  \\ \hline
\multicolumn{1}{|l|}{0.28}            & \multicolumn{1}{l|}{0.21}             & 0.21                                     & \multicolumn{1}{l|}{0.26}            & 0.14                                  \\ \hline
\multicolumn{1}{|l|}{0.34}            & \multicolumn{1}{l|}{0.27}             & 0.30                                     & \multicolumn{1}{l|}{0.51}            & 0.39                                  \\ \hline
\multicolumn{1}{|l|}{0.42}            & \multicolumn{1}{l|}{0.35}             & 0.31                                     & \multicolumn{1}{l|}{0.29}            & 0.17                                  \\ \hline
\multicolumn{1}{|l|}{0.35}            & \multicolumn{1}{l|}{0.28}             & 0.30                                     & \multicolumn{1}{l|}{0.29}            & 0.17                                  \\ \hline
\multicolumn{1}{|l|}{0.31}            & \multicolumn{1}{l|}{0.24}             & 0.33                                     & \multicolumn{1}{l|}{0.29}            & 0.17                                  \\ \hline
\multicolumn{1}{|l|}{0.24}            & \multicolumn{1}{l|}{0.17}             & 0.22                                     & \multicolumn{1}{l|}{0.46}            & 0.35                                  \\ \hline
\multicolumn{1}{|l|}{0.33}            & \multicolumn{1}{l|}{0.26}             & 0.31                                     & \multicolumn{1}{l|}{0.4}             & 0.29                                  \\ \hline
\multicolumn{1}{|l|}{0.32}            & \multicolumn{1}{l|}{0.25}             & 0.36                                     & \multicolumn{1}{l|}{0.26}            & 0.15                                  \\ \hline
\multicolumn{1}{|l|}{0.33}            & \multicolumn{1}{l|}{0.26}             & 0.36                                     & \multicolumn{1}{l|}{0.26}            & 0.15                                  \\ \hline
\multicolumn{1}{|l|}{0.40}            & \multicolumn{1}{l|}{0.34}             & 0.36                                     & \multicolumn{1}{l|}{0.28}            & 0.17                                  \\ \hline
\multicolumn{1}{|l|}{0.31}            & \multicolumn{1}{l|}{0.25}             & 0.34                                     & \multicolumn{1}{l|}{0.27}            & 0.16                                  \\ \hline
\multicolumn{1}{|l|}{0.25}            & \multicolumn{1}{l|}{0.19}             & 0.24                                     & \multicolumn{1}{l|}{0.27}            & 0.16                                  \\ \hline
\multicolumn{1}{|l|}{0.31}            & \multicolumn{1}{l|}{0.25}             & 0.28                                     & \multicolumn{1}{l|}{0.27}            & 0.16                                  \\ \hline
\multicolumn{1}{|l|}{0.21}            & \multicolumn{1}{l|}{0.14}             & 0.19                                     & \multicolumn{1}{l|}{0.26}            & 0.15                                  \\ \hline
\multicolumn{1}{|l|}{0.35}            & \multicolumn{1}{l|}{0.29}             & 0.39                                     & \multicolumn{1}{l|}{0.47}            & 0.36                                  \\ \hline
\multicolumn{1}{|l|}{0.40}            & \multicolumn{1}{l|}{0.33}             & 0.36                                     & \multicolumn{1}{l|}{0.24}            & 0.13                                  \\ \hline
\multicolumn{1}{|l|}{0.37}            & \multicolumn{1}{l|}{0.31}             & 0.35                                     & \multicolumn{1}{l|}{0.43}            & 0.32                                  \\ \hline
\multicolumn{1}{|l|}{0.33}            & \multicolumn{1}{l|}{0.27}             & 0.30                                     & \multicolumn{1}{l|}{0.24}            & 0.13                                  \\ \hline
\multicolumn{1}{|l|}{0.26}            & \multicolumn{1}{l|}{0.20}             & 0.21                                     & \multicolumn{1}{l|}{0.24}            & 0.13                                  \\ \hline
\multicolumn{1}{|l|}{0.29}            & \multicolumn{1}{l|}{0.23}             & 0.35                                     & \multicolumn{1}{l|}{0.3}             & 0.19                                  \\ \hline
\multicolumn{1}{|l|}{0.36}            & \multicolumn{1}{l|}{0.30}             & 0.31                                     & \multicolumn{1}{l|}{0.37}            & 0.26                                  \\ \hline
\multicolumn{1}{|l|}{0.35}            & \multicolumn{1}{l|}{0.29}             & 0.38                                     & \multicolumn{1}{l|}{0.44}            & 0.33                                  \\ \hline
\multicolumn{1}{|l|}{0.38}            & \multicolumn{1}{l|}{0.32}             & 0.40                                     & \multicolumn{1}{l|}{0.35}            & 0.24                                  \\ \hline
\multicolumn{1}{|l|}{0.29}            & \multicolumn{1}{l|}{0.23}             & 0.27                                     & \multicolumn{1}{l|}{0.41}            & 0.3                                   \\ \hline
\multicolumn{1}{|l|}{0.26}            & \multicolumn{1}{l|}{0.21}             & 0.26                                     & \multicolumn{1}{l|}{0.25}            & 0.14                                  \\ \hline
\multicolumn{1}{|l|}{0.30}            & \multicolumn{1}{l|}{0.25}             & 0.32                                     & \multicolumn{1}{l|}{0.44}            & 0.33                                  \\ \hline
\multicolumn{1}{|l|}{0.31}            & \multicolumn{1}{l|}{0.26}             & 0.33                                     & \multicolumn{1}{l|}{0.35}            & 0.24                                  \\ \hline
\multicolumn{1}{|l|}{0.30}            & \multicolumn{1}{l|}{0.25}             & 0.33                                     & \multicolumn{1}{l|}{0.37}            & 0.26                                  \\ \hline
\multicolumn{1}{|l|}{0.26}            & \multicolumn{1}{l|}{0.21}             & 0.26                                     & \multicolumn{1}{l|}{0.39}            & 0.28                                  \\ \hline
\multicolumn{1}{|l|}{0.37}            & \multicolumn{1}{l|}{0.31}             & 0.34                                     & \multicolumn{1}{l|}{0.25}            & 0.14                                  \\ \hline
\multicolumn{1}{|l|}{0.37}            & \multicolumn{1}{l|}{0.31}             & 0.33                                     & \multicolumn{1}{l|}{0.29}            & 0.18                                  \\ \hline
\multicolumn{1}{|l|}{0.29}            & \multicolumn{1}{l|}{0.24}             & 0.29                                     & \multicolumn{1}{l|}{0.29}            & 0.18                                  \\ \hline
\multicolumn{1}{|l|}{0.32}            & \multicolumn{1}{l|}{0.26}             & 0.34                                     & \multicolumn{1}{l|}{0.29}            & 0.18                                  \\ \hline
\multicolumn{1}{|l|}{0.42}            & \multicolumn{1}{l|}{0.37}             & 0.40                                     & \multicolumn{1}{l|}{0.35}            & 0.24                                  \\ \hline
\multicolumn{1}{|l|}{0.32}            & \multicolumn{1}{l|}{0.26}             & 0.27                                     & \multicolumn{1}{l|}{0.3}             & 0.19                                  \\ \hline
\multicolumn{1}{|l|}{0.47}            & \multicolumn{1}{l|}{0.41}             & 0.45                                     & \multicolumn{1}{l|}{0.24}            & 0.13                                  \\ \hline
\end{tabular}
\end{table}

\begin{table}[!ht]
\caption{NIMA}
\begin{tabular}{|lll|ll|}
\hline
\multicolumn{3}{|c|}{\textbf{One Style Text}}                                                       & \multicolumn{2}{c|}{\textbf{Two Style Texts}}           \\ \hline
\multicolumn{1}{|l|}{\textbf{Sem-CS}} & \multicolumn{1}{l|}{\textbf{Gen-Art}} & \textbf{CLIPStyler} & \multicolumn{1}{l|}{\textbf{Sem-CS}} & \textbf{Gen-Art} \\ \hline
\multicolumn{1}{|l|}{5.67}            & \multicolumn{1}{l|}{3.81}             & 4.93                & \multicolumn{1}{l|}{5.59}            & 3.92             \\ \hline
\multicolumn{1}{|l|}{5.85}            & \multicolumn{1}{l|}{4.06}             & 4.50                & \multicolumn{1}{l|}{6.03}            & 4.39             \\ \hline
\multicolumn{1}{|l|}{5.54}            & \multicolumn{1}{l|}{4.04}             & 4.19                & \multicolumn{1}{l|}{5.26}            & 3.62             \\ \hline
\multicolumn{1}{|l|}{4.98}            & \multicolumn{1}{l|}{3.56}             & 3.79                & \multicolumn{1}{l|}{6.17}            & 4.55             \\ \hline
\multicolumn{1}{|l|}{5.51}            & \multicolumn{1}{l|}{4.11}             & 4.66                & \multicolumn{1}{l|}{6.36}            & 4.8              \\ \hline
\multicolumn{1}{|l|}{5.21}            & \multicolumn{1}{l|}{3.82}             & 4.30                & \multicolumn{1}{l|}{6.25}            & 4.76             \\ \hline
\multicolumn{1}{|l|}{5.96}            & \multicolumn{1}{l|}{4.58}             & 5.13                & \multicolumn{1}{l|}{6.53}            & 5.05             \\ \hline
\multicolumn{1}{|l|}{5.94}            & \multicolumn{1}{l|}{4.57}             & 4.03                & \multicolumn{1}{l|}{5.32}            & 3.85             \\ \hline
\multicolumn{1}{|l|}{5.40}            & \multicolumn{1}{l|}{4.10}             & 4.39                & \multicolumn{1}{l|}{5.66}            & 4.24             \\ \hline
\multicolumn{1}{|l|}{5.70}            & \multicolumn{1}{l|}{4.44}             & 5.04                & \multicolumn{1}{l|}{5.59}            & 4.23             \\ \hline
\multicolumn{1}{|l|}{5.58}            & \multicolumn{1}{l|}{4.33}             & 5.36                & \multicolumn{1}{l|}{5.8}             & 4.44             \\ \hline
\multicolumn{1}{|l|}{5.52}            & \multicolumn{1}{l|}{4.29}             & 4.79                & \multicolumn{1}{l|}{5.44}            & 4.11             \\ \hline
\multicolumn{1}{|l|}{6.04}            & \multicolumn{1}{l|}{4.81}             & 4.60                & \multicolumn{1}{l|}{5.91}            & 4.59             \\ \hline
\multicolumn{1}{|l|}{6.17}            & \multicolumn{1}{l|}{4.97}             & 4.74                & \multicolumn{1}{l|}{6.41}            & 5.09             \\ \hline
\multicolumn{1}{|l|}{5.20}            & \multicolumn{1}{l|}{4.00}             & 3.79                & \multicolumn{1}{l|}{5.91}            & 4.61             \\ \hline
\multicolumn{1}{|l|}{5.58}            & \multicolumn{1}{l|}{4.38}             & 5.02                & \multicolumn{1}{l|}{5.98}            & 4.7              \\ \hline
\multicolumn{1}{|l|}{5.54}            & \multicolumn{1}{l|}{4.35}             & 4.98                & \multicolumn{1}{l|}{5.64}            & 4.36             \\ \hline
\multicolumn{1}{|l|}{4.44}            & \multicolumn{1}{l|}{3.25}             & 3.81                & \multicolumn{1}{l|}{5.28}            & 4.02             \\ \hline
\multicolumn{1}{|l|}{5.45}            & \multicolumn{1}{l|}{4.26}             & 4.64                & \multicolumn{1}{l|}{5.61}            & 4.35             \\ \hline
\multicolumn{1}{|l|}{5.33}            & \multicolumn{1}{l|}{4.15}             & 4.45                & \multicolumn{1}{l|}{6.18}            & 4.92             \\ \hline
\multicolumn{1}{|l|}{5.57}            & \multicolumn{1}{l|}{4.39}             & 4.62                & \multicolumn{1}{l|}{5.88}            & 4.63             \\ \hline
\multicolumn{1}{|l|}{5.27}            & \multicolumn{1}{l|}{4.12}             & 4.05                & \multicolumn{1}{l|}{4.79}            & 3.54             \\ \hline
\multicolumn{1}{|l|}{5.66}            & \multicolumn{1}{l|}{4.51}             & 4.84                & \multicolumn{1}{l|}{5.87}            & 4.62             \\ \hline
\multicolumn{1}{|l|}{4.68}            & \multicolumn{1}{l|}{3.53}             & 3.83                & \multicolumn{1}{l|}{5.15}            & 3.91             \\ \hline
\multicolumn{1}{|l|}{5.68}            & \multicolumn{1}{l|}{4.55}             & 4.81                & \multicolumn{1}{l|}{5.63}            & 4.39             \\ \hline
\multicolumn{1}{|l|}{5.18}            & \multicolumn{1}{l|}{4.06}             & 4.30                & \multicolumn{1}{l|}{5.89}            & 4.65             \\ \hline
\multicolumn{1}{|l|}{5.73}            & \multicolumn{1}{l|}{4.61}             & 5.26                & \multicolumn{1}{l|}{5.33}            & 4.1              \\ \hline
\multicolumn{1}{|l|}{5.36}            & \multicolumn{1}{l|}{4.26}             & 4.36                & \multicolumn{1}{l|}{5.65}            & 4.42             \\ \hline
\multicolumn{1}{|l|}{5.28}            & \multicolumn{1}{l|}{4.19}             & 3.90                & \multicolumn{1}{l|}{5.59}            & 4.36             \\ \hline
\multicolumn{1}{|l|}{5.18}            & \multicolumn{1}{l|}{4.12}             & 4.75                & \multicolumn{1}{l|}{5.99}            & 4.77             \\ \hline
\multicolumn{1}{|l|}{5.32}            & \multicolumn{1}{l|}{4.28}             & 4.79                & \multicolumn{1}{l|}{5.85}            & 4.63             \\ \hline
\multicolumn{1}{|l|}{5.09}            & \multicolumn{1}{l|}{4.06}             & 4.71                & \multicolumn{1}{l|}{5.77}            & 4.56             \\ \hline
\multicolumn{1}{|l|}{5.73}            & \multicolumn{1}{l|}{4.71}             & 5.09                & \multicolumn{1}{l|}{5.53}            & 4.33             \\ \hline
\multicolumn{1}{|l|}{5.07}            & \multicolumn{1}{l|}{4.05}             & 4.28                & \multicolumn{1}{l|}{6.24}            & 5.05             \\ \hline
\multicolumn{1}{|l|}{5.63}            & \multicolumn{1}{l|}{4.63}             & 4.32                & \multicolumn{1}{l|}{5.44}            & 4.25             \\ \hline
\multicolumn{1}{|l|}{4.70}            & \multicolumn{1}{l|}{3.70}             & 4.12                & \multicolumn{1}{l|}{4.72}            & 3.53             \\ \hline
\multicolumn{1}{|l|}{5.81}            & \multicolumn{1}{l|}{4.81}             & 4.87                & \multicolumn{1}{l|}{5.93}            & 4.74             \\ \hline
\multicolumn{1}{|l|}{5.39}            & \multicolumn{1}{l|}{4.40}             & 4.40                & \multicolumn{1}{l|}{5.41}            & 4.23             \\ \hline
\multicolumn{1}{|l|}{4.94}            & \multicolumn{1}{l|}{3.96}             & 4.22                & \multicolumn{1}{l|}{4.94}            & 3.77             \\ \hline
\multicolumn{1}{|l|}{5.85}            & \multicolumn{1}{l|}{4.87}             & 4.71                & \multicolumn{1}{l|}{5.85}            & 4.68             \\ \hline
\multicolumn{1}{|l|}{5.98}            & \multicolumn{1}{l|}{5.00}             & 5.42                & \multicolumn{1}{l|}{5.86}            & 4.69             \\ \hline
\multicolumn{1}{|l|}{4.97}            & \multicolumn{1}{l|}{3.99}             & 4.60                & \multicolumn{1}{l|}{5.26}            & 4.09             \\ \hline
\multicolumn{1}{|l|}{5.54}            & \multicolumn{1}{l|}{4.57}             & 4.99                & \multicolumn{1}{l|}{5.72}            & 4.55             \\ \hline
\multicolumn{1}{|l|}{4.89}            & \multicolumn{1}{l|}{3.92}             & 4.91                & \multicolumn{1}{l|}{5.76}            & 4.6              \\ \hline
\multicolumn{1}{|l|}{5.41}            & \multicolumn{1}{l|}{4.45}             & 4.86                & \multicolumn{1}{l|}{4.99}            & 3.83             \\ \hline
\multicolumn{1}{|l|}{6.19}            & \multicolumn{1}{l|}{5.23}             & 5.77                & \multicolumn{1}{l|}{4.67}            & 3.51             \\ \hline
\multicolumn{1}{|l|}{5.23}            & \multicolumn{1}{l|}{4.29}             & 4.14                & \multicolumn{1}{l|}{5.19}            & 4.03             \\ \hline
\multicolumn{1}{|l|}{5.44}            & \multicolumn{1}{l|}{4.50}             & 4.37                & \multicolumn{1}{l|}{5.5}             & 4.34             \\ \hline
\multicolumn{1}{|l|}{5.25}            & \multicolumn{1}{l|}{4.31}             & 4.91                & \multicolumn{1}{l|}{4.63}            & 3.47             \\ \hline
\multicolumn{1}{|l|}{5.20}            & \multicolumn{1}{l|}{4.27}             & 5.40                & \multicolumn{1}{l|}{5.52}            & 4.36             \\ \hline
\multicolumn{1}{|l|}{5.13}            & \multicolumn{1}{l|}{4.20}             & 4.48                & \multicolumn{1}{l|}{4.52}            & 3.37             \\ \hline
\end{tabular}
\end{table}

\begin{table}[!ht]
\begin{tabular}{|lll|ll|}
\hline
\multicolumn{3}{|c|}{\textbf{One Style Text}}                                                       & \multicolumn{2}{c|}{\textbf{Two Style Texts}}           \\ \hline
\multicolumn{1}{|l|}{\textbf{Sem-CS}} & \multicolumn{1}{l|}{\textbf{Gen-Art}} & \textbf{CLIPStyler} & \multicolumn{1}{l|}{\textbf{Sem-CS}} & \textbf{Gen-Art} \\ \hline
\multicolumn{1}{|l|}{5.92}            & \multicolumn{1}{l|}{4.99}             & 5.60                & \multicolumn{1}{l|}{4.69}            & 3.55             \\ \hline
\multicolumn{1}{|l|}{4.03}            & \multicolumn{1}{l|}{3.10}             & 4.05                & \multicolumn{1}{l|}{4.32}            & 3.18             \\ \hline
\multicolumn{1}{|l|}{4.83}            & \multicolumn{1}{l|}{3.90}             & 4.54                & \multicolumn{1}{l|}{5.18}            & 4.04             \\ \hline
\multicolumn{1}{|l|}{4.33}            & \multicolumn{1}{l|}{3.41}             & 4.19                & \multicolumn{1}{l|}{4.81}            & 3.67             \\ \hline
\multicolumn{1}{|l|}{4.26}            & \multicolumn{1}{l|}{3.34}             & 3.68                & \multicolumn{1}{l|}{5.34}            & 4.21             \\ \hline
\multicolumn{1}{|l|}{4.67}            & \multicolumn{1}{l|}{3.76}             & 4.25                & \multicolumn{1}{l|}{5.83}            & 4.7              \\ \hline
\multicolumn{1}{|l|}{4.75}            & \multicolumn{1}{l|}{3.85}             & 4.34                & \multicolumn{1}{l|}{5.77}            & 4.64             \\ \hline
\multicolumn{1}{|l|}{6.05}            & \multicolumn{1}{l|}{5.15}             & 5.88                & \multicolumn{1}{l|}{5.8}             & 4.67             \\ \hline
\multicolumn{1}{|l|}{4.94}            & \multicolumn{1}{l|}{4.04}             & 3.96                & \multicolumn{1}{l|}{4.8}             & 3.68             \\ \hline
\multicolumn{1}{|l|}{5.43}            & \multicolumn{1}{l|}{4.53}             & 4.70                & \multicolumn{1}{l|}{5.46}            & 4.35             \\ \hline
\multicolumn{1}{|l|}{5.10}            & \multicolumn{1}{l|}{4.22}             & 4.72                & \multicolumn{1}{l|}{6.24}            & 5.13             \\ \hline
\multicolumn{1}{|l|}{5.08}            & \multicolumn{1}{l|}{4.21}             & 4.75                & \multicolumn{1}{l|}{5.37}            & 4.27             \\ \hline
\multicolumn{1}{|l|}{5.61}            & \multicolumn{1}{l|}{4.74}             & 4.99                & \multicolumn{1}{l|}{5.66}            & 4.56             \\ \hline
\multicolumn{1}{|l|}{5.84}            & \multicolumn{1}{l|}{4.97}             & 4.74                & \multicolumn{1}{l|}{5.76}            & 4.66             \\ \hline
\multicolumn{1}{|l|}{4.05}            & \multicolumn{1}{l|}{3.19}             & 3.91                & \multicolumn{1}{l|}{5.81}            & 4.71             \\ \hline
\multicolumn{1}{|l|}{6.08}            & \multicolumn{1}{l|}{5.22}             & 5.43                & \multicolumn{1}{l|}{5.31}            & 4.21             \\ \hline
\multicolumn{1}{|l|}{5.01}            & \multicolumn{1}{l|}{4.16}             & 4.73                & \multicolumn{1}{l|}{5.34}            & 4.24             \\ \hline
\multicolumn{1}{|l|}{5.40}            & \multicolumn{1}{l|}{4.56}             & 5.16                & \multicolumn{1}{l|}{4.9}             & 3.81             \\ \hline
\multicolumn{1}{|l|}{4.87}            & \multicolumn{1}{l|}{4.03}             & 4.41                & \multicolumn{1}{l|}{5.25}            & 4.17             \\ \hline
\multicolumn{1}{|l|}{5.43}            & \multicolumn{1}{l|}{4.59}             & 5.16                & \multicolumn{1}{l|}{5.34}            & 4.26             \\ \hline
\multicolumn{1}{|l|}{5.67}            & \multicolumn{1}{l|}{4.83}             & 4.56                & \multicolumn{1}{l|}{5.38}            & 4.3              \\ \hline
\multicolumn{1}{|l|}{5.67}            & \multicolumn{1}{l|}{4.83}             & 5.23                & \multicolumn{1}{l|}{6.06}            & 4.98             \\ \hline
\multicolumn{1}{|l|}{5.21}            & \multicolumn{1}{l|}{4.38}             & 4.17                & \multicolumn{1}{l|}{5.92}            & 4.85             \\ \hline
\multicolumn{1}{|l|}{4.15}            & \multicolumn{1}{l|}{3.32}             & 4.22                & \multicolumn{1}{l|}{5.56}            & 4.49             \\ \hline
\multicolumn{1}{|l|}{5.24}            & \multicolumn{1}{l|}{4.41}             & 3.89                & \multicolumn{1}{l|}{5.51}            & 4.44             \\ \hline
\multicolumn{1}{|l|}{6.04}            & \multicolumn{1}{l|}{5.22}             & 5.58                & \multicolumn{1}{l|}{5.97}            & 4.9              \\ \hline
\multicolumn{1}{|l|}{5.17}            & \multicolumn{1}{l|}{4.35}             & 4.27                & \multicolumn{1}{l|}{5.62}            & 4.56             \\ \hline
\multicolumn{1}{|l|}{6.00}            & \multicolumn{1}{l|}{5.18}             & 4.91                & \multicolumn{1}{l|}{4.83}            & 3.77             \\ \hline
\multicolumn{1}{|l|}{5.44}            & \multicolumn{1}{l|}{4.62}             & 4.71                & \multicolumn{1}{l|}{5.85}            & 4.79             \\ \hline
\multicolumn{1}{|l|}{5.67}            & \multicolumn{1}{l|}{4.85}             & 5.24                & \multicolumn{1}{l|}{4.68}            & 3.62             \\ \hline
\multicolumn{1}{|l|}{5.28}            & \multicolumn{1}{l|}{4.46}             & 4.49                & \multicolumn{1}{l|}{5.94}            & 4.89             \\ \hline
\multicolumn{1}{|l|}{5.32}            & \multicolumn{1}{l|}{4.51}             & 4.94                & \multicolumn{1}{l|}{5.73}            & 4.68             \\ \hline
\multicolumn{1}{|l|}{5.40}            & \multicolumn{1}{l|}{4.58}             & 4.32                & \multicolumn{1}{l|}{5.82}            & 4.77             \\ \hline
\multicolumn{1}{|l|}{5.14}            & \multicolumn{1}{l|}{4.33}             & 4.48                & \multicolumn{1}{l|}{5.32}            & 4.27             \\ \hline
\multicolumn{1}{|l|}{4.58}            & \multicolumn{1}{l|}{3.77}             & 4.57                & \multicolumn{1}{l|}{5.67}            & 4.62             \\ \hline
\multicolumn{1}{|l|}{5.20}            & \multicolumn{1}{l|}{4.40}             & 3.92                & \multicolumn{1}{l|}{5.5}             & 4.45             \\ \hline
\multicolumn{1}{|l|}{5.40}            & \multicolumn{1}{l|}{4.60}             & 4.38                & \multicolumn{1}{l|}{5.22}            & 4.17             \\ \hline
\multicolumn{1}{|l|}{4.99}            & \multicolumn{1}{l|}{4.20}             & 5.01                & \multicolumn{1}{l|}{4.9}             & 3.86             \\ \hline
\multicolumn{1}{|l|}{5.37}            & \multicolumn{1}{l|}{4.58}             & 4.92                & \multicolumn{1}{l|}{5.75}            & 4.71             \\ \hline
\multicolumn{1}{|l|}{5.57}            & \multicolumn{1}{l|}{4.78}             & 4.41                & \multicolumn{1}{l|}{5.72}            & 4.68             \\ \hline
\multicolumn{1}{|l|}{5.68}            & \multicolumn{1}{l|}{4.88}             & 4.63                & \multicolumn{1}{l|}{5.22}            & 4.18             \\ \hline
\multicolumn{1}{|l|}{5.85}            & \multicolumn{1}{l|}{5.05}             & 4.72                & \multicolumn{1}{l|}{5.92}            & 4.89             \\ \hline
\multicolumn{1}{|l|}{4.35}            & \multicolumn{1}{l|}{3.55}             & 3.41                & \multicolumn{1}{l|}{5.66}            & 4.63             \\ \hline
\multicolumn{1}{|l|}{5.79}            & \multicolumn{1}{l|}{5.00}             & 4.85                & \multicolumn{1}{l|}{5.28}            & 4.25             \\ \hline
\multicolumn{1}{|l|}{5.77}            & \multicolumn{1}{l|}{4.99}             & 4.71                & \multicolumn{1}{l|}{5.06}            & 4.03             \\ \hline
\multicolumn{1}{|l|}{5.79}            & \multicolumn{1}{l|}{5.01}             & 4.06                & \multicolumn{1}{l|}{5.27}            & 4.24             \\ \hline
\multicolumn{1}{|l|}{5.22}            & \multicolumn{1}{l|}{4.44}             & 4.80                & \multicolumn{1}{l|}{5.23}            & 4.21             \\ \hline
\multicolumn{1}{|l|}{4.72}            & \multicolumn{1}{l|}{3.95}             & 4.66                & \multicolumn{1}{l|}{5.66}            & 4.64             \\ \hline
\multicolumn{1}{|l|}{5.69}            & \multicolumn{1}{l|}{4.92}             & 5.23                & \multicolumn{1}{l|}{5.28}            & 4.26             \\ \hline
\end{tabular}
\end{table}
\vfill\pagebreak
\end{document}